\documentclass{article}
\usepackage[a4paper, margin=1in]{geometry}

\usepackage{hyperref}
\usepackage{amsmath}
\usepackage{color}
\usepackage{xcolor}
\usepackage{graphicx}
\usepackage{multirow}
\usepackage{xspace}
\usepackage{url}
\usepackage[defaultlines=4,all]{nowidow}
\usepackage{soul}
\usepackage[round]{natbib}

\definecolor{darkblue}{rgb}{0, 0, 0.6}
\hypersetup{colorlinks=true,citecolor=darkblue, linkcolor=darkblue, urlcolor=darkblue}

\newcommand{\papertitle}{Native Language Identification using Stacked Generalization}

\begin{document}
\title{\papertitle}

\author{
  Shervin Malmasi\\
  Harvard Medical School\\
  \texttt{shervin.malmasi@mq.edu.au}
  \and
  Mark Dras\\
  Macquarie University\\
  \texttt{mark.dras@mq.edu.au}
}

\date{March 2017}

\newcommand{\lex}[1]{\textit{#1}}
\newcommand{\cfgrule}[1]{\texttt{#1}}
\newcommand{\posgram}[1]{\lstinline!#1!}
\newcommand{\lgf}[1]{\textit{#1}}

\newcommand\tfl{{\sc Toefl$11$}\xspace}
\newcommand{\tflb}{\sc\bf{\textsc{Toefl$\mathbf{11}$}}\xspace}
\newcommand\tfltrain{{\sc Toefl$11$-Train}\xspace}
\newcommand\tfldev{{\sc Toefl$11$-Dev}\xspace}
\newcommand\tfltest{{\sc Toefl$11$-Test}\xspace}
\newcommand\tfltdev{{\sc Toefl$11$-TrainDev}\xspace}
\newcommand{\efc}{\textsc{EfCamDat}\xspace}

\DeclareRobustCommand{\eg}{\textit{e.g.}\@\xspace}
\DeclareRobustCommand{\ie}{\textit{i.e.}\@\xspace}
\DeclareRobustCommand{\vav}{\textit{vis-\`a-vis}\@\xspace}

\makeatletter
\DeclareRobustCommand{\etc}{
    \@ifnextchar{.}%
        {etc}%
        {etc.\@\xspace}%
}
\makeatother

\DeclareRobustCommand{\ng}{$n$-gram\@\xspace}
\DeclareRobustCommand{\ngs}{$n$-grams\@\xspace}
\DeclareRobustCommand{\ngb}{$\mathbf{n}$\bf{-gram}\@\xspace}
\DeclareRobustCommand{\ngsb}{$\mathbf{n}$\bf{-grams}\@\xspace}

\newcommand\hpt{\hbox{$\cdot$}}

\let\cite=\citep
\maketitle

\begin{abstract}
Ensemble methods using multiple classifiers have proven to be the most successful approach for the task of Native Language Identification (NLI), achieving the current state of the art.
However, a systematic examination of ensemble methods for NLI has yet to be conducted.
Additionally, deeper ensemble architectures such as classifier stacking have not been closely evaluated.
We present a set of experiments using three ensemble-based models, testing each with multiple configurations and algorithms. This includes a rigorous application of meta-classification models for NLI, achieving state-of-the-art results on three datasets from different languages.
We also present the first use of statistical significance testing for comparing NLI systems, showing that our results are significantly better than the previous state of the art. We make available a collection of test set predictions to facilitate future statistical tests.

\end{abstract}

\section{Introduction}

Native Language Identification (NLI) is the task of identifying a writer's native language (L1) based only on their writings in a second language (the L2).
NLI works by identifying language use patterns that are common to groups of
speakers of the same native language. This process is
underpinned by the presupposition that an author's L1 disposes
them towards certain language production patterns in their L2, as
influenced by their mother tongue. This relates to cross-linguistic influence (CLI), a key topic in the field of Second Language Acquisition (SLA) that analyzes transfer effects from the L1 on later learned languages \cite{ortega:2009}.

It has been noted in the linguistics literature since 
the 1950s that speakers of particular languages
have characteristic production patterns
when writing in a second language.
This language transfer phenomenon has been investigated independently
in various fields from different perspectives, including
qualitative research in SLA and 
recently via predictive models in NLP \cite{jarvis2012approaching}.
Recently this has motivated studies in NLI, a subtype of text classification where the goal is to determine the
native language  of an author using texts they have written in a second language or L2 \cite{tetreault-etal:2013:BEA}.

The motivations for NLI are manifold. 
Such techniques can help SLA researchers identify important L1-specific learning and teaching issues. In turn, the identification of such issues can enable researchers to develop pedagogical material that takes into consideration a learner's L1 and addresses them.
It can also be applied in a forensic context, for example,
to glean information about the discriminant L1 cues in an anonymous text.

NLI is most commonly framed as a multi-class supervised classification task.
Researchers have experimented with a range of machine learning algorithms, with Support Vector Machines having found the most success.
However, some of the most successful approaches have made use of classifier ensemble methods to further improve performance on this task.
This is a trend that has become apparent in recent work on this task, as we'll outline in \S\ref{sec:relwork}. In fact, all recent state-of-the-art systems have relied on some form of multiple classifier system.

However, a thorough examination of ensemble methods for NLI --- one empirically comparing different architectures and algorithms --- has yet to be conducted.
Additionally, more sophisticated ensemble architectures, such as stacked generalization (classifier stacking), have not been closely evaluated.
This meta-classification approach is an advanced method that has proven to be effective in many classification tasks;
its systematic application could improve the state of the art in NLI.

This has links to the idea of adding layers to increase power in neural network-based deep learning, which has come to be an important approach in NLP over the last couple of years \citep{manning:2015:CL}; \citet{eldan-shamir:2016:JMLR} note that ``Overwhelming empirical evidence as well as intuition indicates that having depth in the neural network is indeed important".  Deep neural networks can in fact be seen as layered classifiers \citep{goldberg:2015:arxiv}, and ensemble methods as an alternative way of adding power via additional layers.  In this article we look just at ensemble methods: deep learning has not yet produced state-of-the-art results on related tasks
\citep{dsl2016},\footnote{Traditional text classification methods substantially outperformed all deep learning approaches in the 2016 DSL Shared Task.}
and our goal is to understand what it is that has made ensemble methods to date in NLI so successful.

The primary focus of the present work is to address this gap by presenting a comprehensive and rigorous examination of how ensemble methods can be applied for NLI. We aim to examine several different ensemble and meta-classification architectures, each of which can utilize different configurations and algorithms.

Furthermore, previous ensemble methods have not been tested on different datasets, making the generalizability of these models for NLI unclear. Ideally, the same method should be tested across multiple corpora to assess its validity.
When working on a common dataset, authors should also aim to compare the performance of their methods directly.
To this end, we also apply our methods to three datasets to evaluate their generalizability.
NLI methods have been recently applied to different languages and we believe that this type of multilingual evaluation is an important trend for future NLI research.
Our chosen datasets therefore include the most commonly used English NLI corpus as well as more recently used Chinese and Norwegian corpora.

The final aspect of this work deals with evaluation, which in NLI work thus far has relied mostly on direct comparisons between the reported accuracies of various systems and their relative differences. However, as the reported performance continues to rise, it becomes more important to compare and interpret these results objectively. Although statistical methods can facilitate such an objective interpretation and comparison between systems, they have not been used in NLI, for reasons which we will outline later.
Consequently, the final objective of this study is to apply statistical methods for comparing different approaches.
We not only compare our results against those previously reported, but also conduct statistical significance testing against other state-of-the-art NLI systems, something which has not been performed to date.

To summarize, the principal aims of the present study are to:

\begin{enumerate}
\item Apply several advanced ensemble combination methods to NLI and evaluate their performance against previously used ensemble methods.

\item Evaluate the use of meta-classifiers for NLI, applying different feature representations and a range of learning algorithms.

\item Compare the performance of these methods to previous results
and
assess the methods on different languages/datasets.

\item Investigate the use of statistical testing for comparing NLI systems.
\end{enumerate}

The remainder of this paper is organized as follows.
In \S\ref{sec:relwork} we introduce ensemble classification and recap previous work in NLI. Our data is introduced in \S\ref{sec:data}, followed by our experimental setup in \S\ref{sec:method}, and our classification features in \S\ref{sec:features}.
Our ensemble-based models are detailed in \S\ref{sec:models} and experimental results are reported in \S\ref{sec:results}. We then conclude with a discussion in \S\ref{sec:discussion}.

\section{Related Work}
\label{sec:relwork}

This work draws on two broad areas of research: ensemble-based classification methods and work in NLI.

\subsection{Ensemble Classifiers}
\label{sec:ensemble-relwork}

Classifier ensembles are a way of combining different classifiers or experts with the goal of improving overall accuracy through enhanced decision making.
Instead of relying on decisions by a single expert, they attempt to reach a decision by utilizing the collective input from a committee of experts.

They have been applied to a wide range of real-world problems and shown to achieve better results compared to single-classifier methods \citep{oza2008classifier}.
Through aggregating the outputs of multiple classifiers in some way, their outputs are generally considered to be more robust.
Ensemble methods continue to receive increasing attention from investigators and remain a focus of machine learning research \citep{wozniak2014survey,kuncheva2014weighted}.

Such ensemble-based systems often use a parallel architecture, where the classifiers are run independently and their outputs are aggregated using a fusion method.
The specifics of how such systems work will be detailed in section \ref{sec:models}.

They have been applied to various classification tasks with good results.
Not surprisingly, researchers have attempted to use them for improving the performance of NLI, as we discuss in the next section.

\subsection{Native Language Identification}

NLI work has been growing in recent years,
using a wide range of syntactic and more recently, lexical features to distinguish the L1.
A detailed review of NLI methods is omitted here for reasons of space, but a thorough exposition is presented in the report from the very first NLI Shared Task that was held in 2013 \cite{tetreault-etal:2013:BEA}.

Most English NLI work has been done using two corpora.
The \textit{International Corpus of Learner English} \cite{Granger:2009}
was widely used until recently, despite its shortcomings\footnote{The issues exist as the corpus was not designed specifically for NLI.} being widely noted \cite{Brooke:2012}.
More recently, \tfl, the first corpus designed for NLI was released \cite{blanchard-tetreault-higgins-cahill-chodorow:2013:TOEFL11-RR}.
While it is the largest NLI dataset available, 
it only contains
argumentative essays,
limiting analyses to this genre.

Research has also expanded to use non-English learner corpora \cite{malmasi:2014:anli,malmasi:2014:fnli}.
Recently, 
\citet{malmasi:2014:cnli} introduced the Jinan Chinese Learner Corpus \cite{wang-malmasi-huang:2015} for NLI 
and their results indicate that feature performance may be similar across corpora and even L1-L2 pairs.
Similarly, \citet{malmasi:2015:nnli} also proposed using the ASK corpus to conduct NLI research using L2 Norwegian data.
In this study we make use of three of these aforementioned corpora: \tfl, JCLC and ASK; detailed descriptions will be provided in \S\ref{sec:data}.

As mentioned earlier, some of the most successful approaches to NLI have used ensemble learning methods.
We now present an overview of this ensemble-based NLI research.

\citet{tetreault-EtAl:2012:PAPERS} were the first to propose the use of classifier ensembles for NLI and performed a comprehensive evaluation of the feature types used until that point. In their study they used an ensemble of logistic regression learners using a wide range of features that included character and word $n$-grams, function words, parts of speech, spelling errors and writing quality markers. With regard to syntactic features, they also investigated the use of Tree Substitution Grammars and dependency features extracted using the Stanford parser. Furthermore, they also proposed using language models for this task and in their system used language model perplexity scores based on lexical $5$-grams from each language in the corpus. The set of features used here was the largest of any NLI study to date. With this system, the authors reported state of the art accuracies of $90.1\%$ and $80.9\%$ on the ICLE and \tfl corpora, respectively.
\citet{tetreault-EtAl:2012:PAPERS} also conducted cross-corpus evaluation, using the 7 common L1 classes between the ICLE and \tfl corpora. Training on the ICLE data, they report an accuracy of 26.6\%.

The very first shared task focusing on Native Language Identification was held in 2013, bringing further focus, interest and attention to the field. The NLI Shared Task 2013\footnote{\url{https://sites.google.com/site/nlisharedtask2013/home}} was co-located with the eighth instalment of the Building Educational Applications Workshop at NAACL-HLT 2013.

The competition attracted entries from 29 teams. 
The winning entry for the shared task was that of \citet{jarvis-bestgen-pepper:2013:BEA8}, with an accuracy of 83.6\%. The features used in this system are $n$-grams of words, parts-of-speech as well as lemmas. In addition to normalizing each text to unit length, the authors also applied a log-entropy weighting schema to the normalized values, which clearly improved the accuracy of the model. An L2-regularized SVM classifier was used to create a single-model system. Furthermore, the authors employed their own procedure for optimizing the cost parameter (C) of the SVM. While they did not use a great number of features or introduce any new features for this task, we posit that their use of weighting schema and hyperparameter optimization gave their system an edge over their competitors, the majority of whom did not employ these techniques.

A notable trend among the other entries was the use of ensemble-based systems, which have been shown to achieve better results over systems based on single models. We will now briefly review the systems that took this approach.

\citet{gyawali-ramirez-solorio:2013:BEA8} 
utilized lexical and syntactic features based on $n$-grams of characters, words and part-of-speech tags (using both the Penn TreeBank and Universal Parts Of Speech tagsets), along with perplexity values of character $n$-grams to build four different models. These models were combined using a voting-based ensemble of SVM classifiers. Features values were weighted using the TF-IDF scheme. In particular, the authors set out to investigate whether a more coarse grained POS tagset would be useful for NLI. They explore the use of the Universal POS tagset which has 12 POS categories in the NLI shared task and compare the results with the fine-grained Penn TreeBank (PTB) tagset that includes 36 POS categories. The highest accuracy of their system in the shared task is 74.8\%, achieved by combining all features into an ensemble. The authors found that the use of coarse grained Universal POS tags as features generalizes the syntactic information and reduces the discriminative power of the feature that comes from the fine granularity of the $n$-grams. For example, the PTB tagset distinguishes verbs into six distinct categories while the Universal POS tagset only has a single category for that grammatical class.

In the system designed by \citet{cimino-EtAl:2013:BEA8} the authors use a wide set of general purpose features that are designed to be portable across languages, domains and tasks. This set includes features that are lexical (sentence length, document length, type/token ration, character and word $n$-grams), morpho-syntactic (coarse and fine-grained part-of-speech tag $n$-grams) and syntactic (parse tree and dependency-based features). They report that they found distributional differences across the L1s for many of these features, including average word and sentence lengths.  However, we note that many of these differences are not of a large magnitude, and the authors did not run any statistical tests to measure the significance levels of these differences.  Using this feature set, they experiment with a single-classifier system as well as classifier ensembles, using SVM and Maximum Entropy classifiers. In their ensemble, they experiment with using a majority voting system as well as a meta-classifier approach. The authors report that the ensemble methods outperform all single-classifier systems (by around 2\%), and their best performance of 77.9\% is provided by the meta-classifier system which used linear SVM and MaxEnt as the component classifiers and combined the results using a polynomial kernel SVM classifier. While the set of features used in this experiment is not widely different to other reported NLI research, their use of a meta-classifier is an interesting approach that warrants further study.

In their system, \citet{goutte-leger-carpuat:2013:BEA8} used character, word and part-of-speech $n$-grams along with syntactic dependencies. They used an ensemble of SVM classifiers trained on each feature space, using a majority vote combiner method. To represent the feature values, they use two value normalization methods based on TF-IDF and cosine normalization. Their best entry achieved an accuracy of 81.8\%, higher than many systems using the same standard features and more, demonstrating the effectiveness of using ensemble classifiers and appropriate feature value representation. The authors, like many others, also note that lexical features provided the best performance  for a single feature in their system, but that this can be boosted by combining multiple predictors.

The MITRE system \citep{henderson-EtAl:2013:BEA8} is another highly lexicalized system where the primary features used are word, part-of-speech and character $n$-grams. In this system, these features are used by independent classifiers (logistic regression, Winnow2 and language models) whose output is then combined into a final prediction using a Na{\"i}ve Bayes model. Their best performing ensemble was 82.6\% accurate in the shared task and the authors emphasize the value of ensemble methods that combine independent systems. Furthermore, the authors also optimized the parameters of their Naive Bayes model using a grid search over the development data.

\citet{hladka-holub-kriz:2013:BEA8} developed an ensemble classifier system using some standard features (lemma, word and part-of-speech $n$-grams, word skipgrams) with SVM classifiers. They obtained an accuracy of 72.5\% in the shared task. They found that their ensemble, which is based on majority voting, outperformed other methods of combining the features. This is yet another piece of evidence pointing to the utility of using ensemble systems for NLI.

Another system that utilizes an ensemble is that of \citet{bykh-EtAl:2013:BEA8}, where they used a probability-based ensemble.
They use a set of 16 features, including recurring word-based $n$-grams, recurring Open Class POS (OCPOS) $n$-grams, dependencies, trees and lemmas. To combine the different feature types, they explored combining all feature into a single vector and also ensembles of SVM classifiers (each trained on a single feature type). Their best shared task performance of 82.2\% was achieved using an ensemble with all of their features. Their analysis shows that recurring word-based $n$-grams are the best performing single feature type, once again demonstrating the relevance and significant of lexical features in NLI.

Following the shared task, \citet{bykh:2014} further explored the use of lexicalized and non-lexicalized phrase structure rules for NLI. They show that the inclusion of lexicalized production rules (\ie preterminal nodes and terminals) provides improved results.
In addition to the standard normalized frequency and binary feature representations they also propose two new representations based on a ``variationist sociolinguistic" perspective.
Although they show that these representations outperform the normalized frequency approach, they do not compare this to other representations which have been shown to improve NLI accuracy, such as TF-IDF.
They combine their lexicalized production rules feature with additional surface \ng features in a tuned and optimized ensemble, reporting an accuracy of 84.82\% on the \tfltest set.

\citet{ionescu:2014:nli} extend the previous work of \citet{popescu-ionescu:2013:BEA8} which used string kernels to perform NLI using only character \ng features.
One improvement here is that several string kernels are combined through multiple kernel learning.
Although this approach is not based on the types of ensembles we use here, it is similar in the sense that it attempts to combine multiple learners.
The authors also perform parameter tuning to select the optimal settings for their system.
They report an accuracy of $85.3\%$ on the \tfltest set, $1.7\%$ higher than the winning shared task system.
Recently they expanded their approach with additional experiments \citep{ionescu2016string}, although they did not achieve further improvements on \tfltest.
One shortcoming of this approach is that they do not present a single model that achieves best performance on the different sets; different parameters are used to achieved the best results for each set.
This can be theoretically unsatisfying since it is possible that the different parameters could be overfitting the sets.
Ideally, a single model with fixed parameters would obtain the best result across all sets.

In sum, we can see that ensemble-based approaches have yielded some of the most successful results in this field. However, we also believe that it is possible to employ ensemble models that are even more sophisticated, leading to improved results. This is the key research question being investigated by the present study.

\section{Data}
\label{sec:data}
We now introduce the three datasets used in this study.
One of the goals of this study is to assess the generalizability of the methods and results across datasets, and this requires us to use multiple corpora.
They have all been used in previous NLI work and cover different (second) languages: English, Chinese and Norwegian.

\subsection{The \textsc{TOEFL11} Corpus}
\label{sec:data-t11}

The \tfl corpus \citep{blanchard-tetreault-higgins-cahill-chodorow:2013:TOEFL11-RR} --- also known as the \textit{ETS Corpus of Non-Native Written English} --- is the first dataset designed specifically for the task of NLI and developed with the aim of addressing the above-mentioned deficiencies of other previously used corpora. By providing a common set of
L1s and evaluation standards, the authors set out to facilitate the direct comparison of approaches and methodologies.

Furthermore, as all of the texts were collected through the Educational Testing Service's electronic test delivery system, this ensures that all of the data files are encoded and stored in a consistent manner.\footnote{The essays are distributed as UTF-8 encoded text files.} The corpus is available through the the Linguistic Data Consortium.\footnote{https://catalog.ldc.upenn.edu/LDC2014T06}

It consists of 12,100 learner texts from speakers of 11 different languages. The texts are independent task essays written in response to eight different prompts, and were collected in the process of administering the Test of English as a Foreign Language (TOEFL\textregistered) between 2006-2007.
The texts are divided into specific training (\tfltrain), development (\tfldev) and test (\tfltest) sets.
It is also common to combine the training and development sets for training, which we refer to as \tfltdev.

\begin{figure}
\centering
\includegraphics[width=1\textwidth]{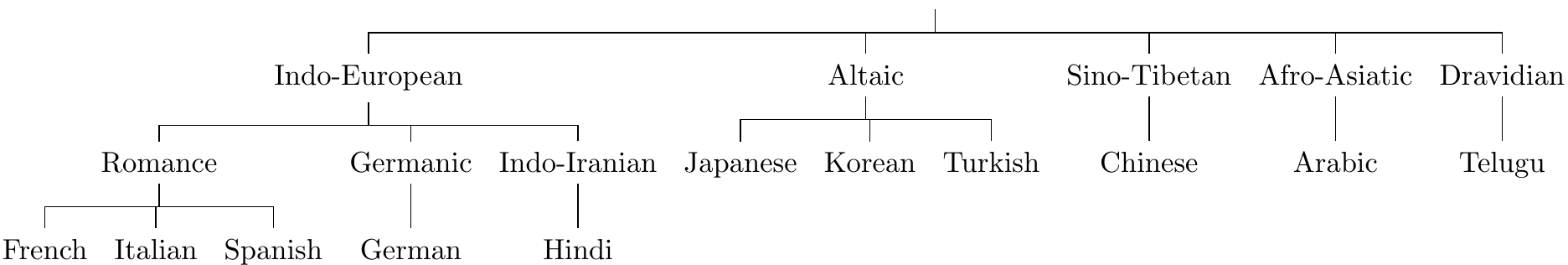}
\caption[Languages in the \tfl corpus.]{Language families in the \tfl corpus. The languages were selected to represent different families, but to also have several from within the same families. Diagram reproduced from \citet{blanchard-tetreault-higgins-cahill-chodorow:2013:TOEFL11-RR}.}
\label{fig:TOEFL11-languages}
\end{figure}

The 11 L1s are Arabic, Chinese, French, German, Hindi, Italian, Japanese, Korean, Spanish, Telugu and Turkish.
This selection ensures that there are L1s from diverse language families, but also several from within certain families.
The L1s and their language families are shown in Figure~\ref{fig:TOEFL11-languages}. 

\begin{figure}
\centering
\includegraphics[width=1\textwidth]{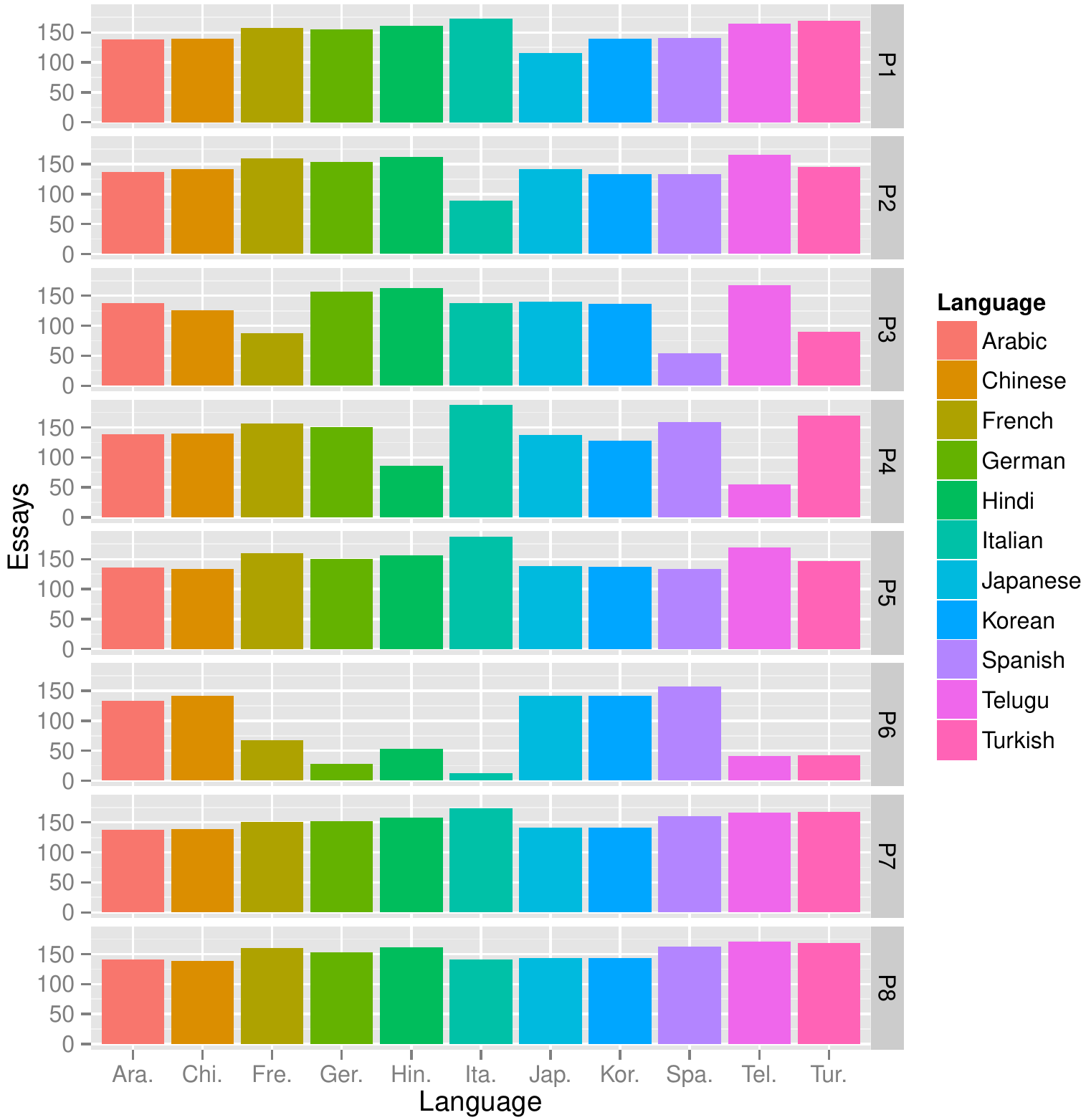}
\caption[Distribution of essay prompts by L1 in the \tfl corpus]{A plot of how the eight topic prompts are distributed across the L1 groups in the \tfl corpus. Prompts are labelled P1--P8. Figure reproduced from \citet{blanchard-tetreault-higgins-cahill-chodorow:2013:TOEFL11-RR}.}
\label{fig:TOEFL11-prompts}
\end{figure}

This dataset was designed specifically for NLI and the authors attempted to balance the texts by topic and native language.
There are a total of eight essay prompts in the corpus, with the prompts setting each essay's topic or theme.
Although they were not able to create a perfectly balanced corpus, the distribution of topics across L1s is very even.
This distribution of essay prompts by L1 is shown in Figure~\ref{fig:TOEFL11-prompts}.

\subsection{The ASK Corpus}
\label{sec:ask-corpus}

In this study we use data from the ASK Corpus (\textit{Andrespr\r{a}kskorpus}, Second Language Corpus).
The ASK Corpus \cite{ASK:2013,ASK,tenfjord2006hows} is a learner corpus composed of the writings of learners of Norwegian. These texts are essays written as part of a test of Norwegian as a second language. Each text also includes additional metadata about the author such as age or native language. An advantage of this corpus is that all the texts have been collected under the same conditions and time limits.
The corpus also contains a control subcorpus of texts written by native Norwegians under the same test conditions. The corpus also includes error codes and corrections, although we do not make use of this information here.

There are a total of $1{,}700$ essays written by learners of Norwegian as a second language with ten different first languages: German, Dutch, English, Spanish, Russian, Polish, Bosnian-Croatian-Serbian, Albanian, Vietnamese and Somali.
The essays are written on a number of different topics, but these topics are not balanced across the L1s.

Detailed word level annotations (lemma, POS tag and grammatical function) have been first obtained automatically using the Oslo-Bergen tagger. These annotations have then been manually post-edited by human annotators since the tagger's performance can be substantially degraded due to orthographic, syntactic and morphological learner errors. These manual corrections can deal with issues such as unknown vocabulary or wrongly disambiguated words. 

Unlike for TOEFL11, we generate artificial essays here, to mitigate the effects of imbalance in topics and proficiency.
Manufacturing documents in this manner has a number of positive impacts.
Firstly, it ensures that all documents are similar and comparable in length. If the data are being used to classify documents from another source, instead of cross-validation, the generation parameters could be changed so that the training set is similar to the test set in terms of length.
Secondly, the random sampling used here means that the texts created for each class are a mix of different authorship styles, proficiencies and topics. 

In this work we extracted $750$k tokens of text from the ASK corpus in the form of individual sentences.
Following a similar methodology to that of \cite{Brooke:2011}, we randomly select and combine the sentences from the same L1 to generate texts of approximately $300$ tokens on average, creating a set of documents suitable for NLI.

More specifically, the dataset composed of artificial documents is generated as follows.
For each class, all the available texts are processed and the individual sentences from these texts are placed into a single pool. 
Once this pool has been created, we begin the process of generating artificial documents.

For each artificial text to be generated, its required minimum length is first determined by randomly picking a value within a pre-specified range $[M, N]$. This chosen value represents the minimum number of tokens or characters that are required to create a new document. By specifying this range parameter, instead of a single fixed value, we can create an artificial dataset where there is still some reasonable (and controlled) amount of variance in length between texts.

Sentences from the pool are then randomly allocated to the document until its length exceeds the required minimum value. The document is then considered complete; it is added to the new dataset and we proceed to generate another.
It should also be noted that the document length may exceed the upper bound of the range parameter, depending on the length of the final sentence that crosses the minimum threshold.
The sampling of sentences from the pool is done without replacement.

This process continues until there are insufficient sentences to create any more documents. The sentences remaining in the pool are then discarded. This procedure is performed for every class in the original dataset and yields a new dataset of artificial documents.

\begin{figure}
\begin{center}
\includegraphics[width=0.75\textwidth]{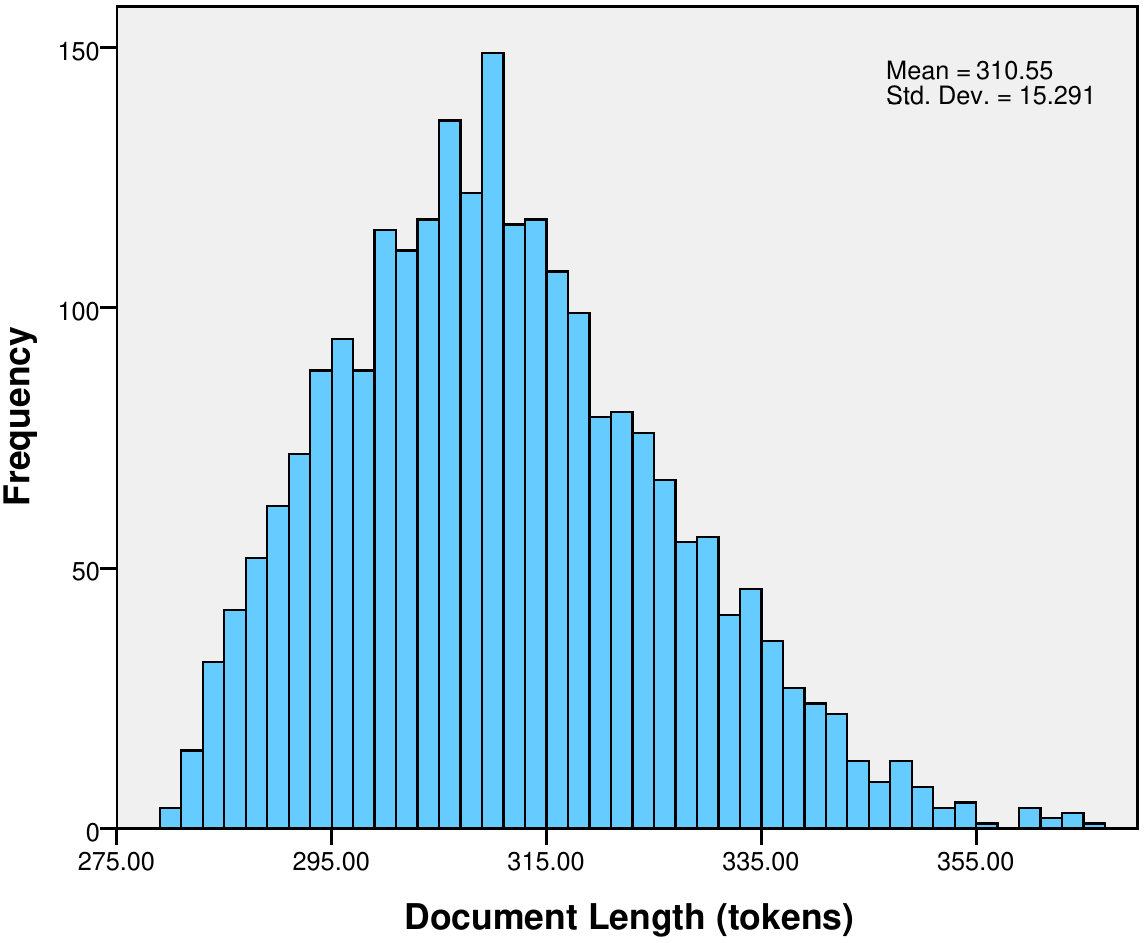}
\end{center}
\caption{A histogram of the number of tokens per document in the dataset that we generated.}
\label{fig:nor-histogram}
\end{figure}

The $10$ native languages and the number of texts generated per class are listed in Table \ref{table:nor-data}. In addition to these we also generate $250$ control texts written by natives.
A histogram of the number of tokens per document is shown in Figure \ref{fig:nor-histogram}. The documents have an average length of $311$ tokens with a standard deviation of $15$ tokens.

\begin{table}
\caption{The $10$ L1 classes included in the Norwegian NLI dataset and the number of texts we generated for each class.}
\label{table:nor-data}
\centering
\tabcolsep=0.15cm
\begin{tabular}{lr}
\hline
Native Language & Documents \\
\hline
Albanian 	& $121$ \\
Dutch		& $254$ \\
English 	& $273$ \\
German 		& $280$ \\
Polish		& $281$ \\
Russian 	& $257$ \\
Serbian 	& $259$ \\
Somali	 	& $90$ \\
Spanish  	& $243$ \\
Vietnamese	& $100$ \\
\hline
\textbf{Total} 	 & $2{,}158$ \\
\hline
\end{tabular}
\end{table}

\subsubsection{Part-of-Speech Tagset}
The ASK corpus uses the Oslo-Bergen tagset\footnote{\url{http://tekstlab.uio.no/obt-ny/english/tagset.html}} which has been developed based on the Norwegian Reference Grammar \cite{faarlund1997norsk}.

Here each POS tag is composed of a set of constituent morphosyntactic tags. For example, the tag \texttt{subst-appell-mask-ub-fl} signifies that the token has the categories ``noun common masculine indefinite plural". Similarly, the tags \texttt{verb-imp} and \texttt{verb-pres} refer to imperative and present tense verbs, respectively.

Given its many morphosyntactic markers and detailed categories, the ASK dataset has a rich tagset with over $300$ unique tags.

\subsection{The Jinan Chinese Learner Corpus}
\label{sec:jclc}

Growing interest has led to the recent development of the Jinan Chinese Learner Corpus \citep{wang-malmasi-huang:2015}, the first large-scale corpus of L2 Chinese consisting of university student essays.
Learners from $59$ countries are represented and proficiency levels are sampled representatively across beginner, intermediate and advanced levels. However, 
texts by learners from other Asian countries are disproportionately represented, with this likely being due to geographical proximity and links to China.

For this work we extracted $3{.}75$ million tokens of text from the JCLC in the form of individual sentences.\footnote{Full texts are not made available, only individual sentences with the relevant metadata (proficiency/nationality).}
Following the methodology described in the previous section, we combine the sentences from the same L1 to generate texts of 600 tokens on average,\footnote{A single Chinese character is considered a token.}
creating a set of documents suitable for NLI.

\begin{table}
\centering
\caption{The $11$ L1 classes included in the Chinese NLI dataset and the number of texts we generated for each class.}
\label{tab:jclc-data}
\tabcolsep=0.15cm
\begin{tabular}{lr}
\hline
Native Language & Documents \\
\hline
Burmese 		& 349	\\
Filipino 		& 415	\\	
Indonesian		& 402	\\	
Japanese	 	& 180	\\
Khmer 			& 294	\\	
Korean			& 330	\\
Laotian 		& 366	\\
Mongolian 		& 101	\\
Spanish			& 112	\\
Thai 			& 400	\\	
Vietnamese		& 267	\\
\hline
\textbf{Total} 	 & $3{,}216$ \\
\hline
\end{tabular}
\end{table}

Although there are over 50 L1s available in the corpus, we choose the top 11 languages, shown in Table \ref{tab:jclc-data}, to use in our experiments. This is due to two considerations. First, while many L1s are represented in the corpus, most have relatively few texts. Choosing the top 11 classes allows us to have a large number of classes and also ensure that there is sufficient data per-class. 
Secondly, this is the same number of classes used in the NLI 2013 shared task, enabling us to draw cross-language comparisons with the shared task results.

\section{Experimental Setup}
\label{sec:method}

In this study we employ a supervised multi-class classification approach.
The learner texts are organized into classes according to the author's L1 and these documents are used for training and testing in our experiments.
In this section we describe our experimental methodology, including evaluation and the algorithms we use.
Our classification features will be described in \S\ref{sec:features} and the three classification models we create using these algorithms and features are then described in \S\ref{sec:models}.

\subsection{Evaluation}
\label{sec:evaluation}

In the same manner as many previous NLI studies and also the NLI 2013 shared task, we report our results as classification accuracy under $k$-fold cross-validation, with $k = 10$. In recent years this has become a \textit{de facto} standard for reporting NLI results.
For creating our folds, we employ stratified cross-validation which aims to ensure that the proportion of classes within each partition is equal \citep{kohavi:1995}. 
For \tfl we also test on the standard test set, which we call \tfltest.

We use a random baseline and a majority class baseline for comparison purposes, using these to determine the lower bounds for accuracy.
Oracles, which we describe in \S\ref{sec:oracle}, will also be used to estimate a potential upper bound for accuracy. We use them to assess how close our models are to achieving optimal performance.
Additionally, we also compare our results against those reported in previous work.
These were described earlier in~\S\ref{sec:relwork}

\subsection{Classification Algorithms}
\label{sec:learners}

In this section we briefly describe the learning algorithms used in our experiments.
All of these learners will be evaluated as meta-classifiers, but only SVMs will be used as base learners, for reasons we outline below (\S\ref{sec:svm}).
Although a thorough exposition of the methods is beyond the scope of this work, key references are provided for the interested reader.

\subsubsection{Linear Support Vector Machine}
\label{sec:svm}
Linear Support Vector Machine (SVM) classifiers are a highly robust supervised classification method that has proven to be very effective for text classification \citep{joachims1998}.
Strong theoretical and empirical arguments have been made for the utility of SVMs for text classification, showing that their capabilities are well suited for several properties of text data, including extremely large feature spaces, very high sparsity and few irrelevant features.

With regards to NLI, post-hoc analysis of the shared task results  revealed that many of the top systems, including the winning entry, followed two broad patterns: they used SVM classifiers along with frequency-based feature values.
Given the better performance of SVM-based NLI systems, we used this learner to generate all of the base classifiers in this study.

SVMs are inherently binary classifiers and a common way to adapt them for multi-class problems is through a one-vs-all (OVA) approach, also known as a one-vs-rest (OVR) approach. Another common alternative is a one-vs-one (OVO) method that builds $\frac{N(N-1)}{2}$ binary classifiers for all pairwise combinations.
It has been found that the OVR approach works best in NLI \citep{brooke-hirst:2012:PAPERS}, and our experiments confirmed this and we therefore adopt this approach.
Additionally, an SVM is a margin-based classifier and does not output probability estimates for each class label, although there are additional methods to map the outputs to probabilities \citep{platt1999probabilistic}.

\subsubsection{RBF-Kernel Support Vector Machine}
SVMs with a Radial Basis Function (RBF) kernel are also popular for data points that are not linearly separable.
This is because the kernel maps the data points in a non-linear manner, allowing for more flexible decision boundaries \cite{hsu2003practical}. It should also be noted that this flexibility increases the risk of overfitting.

Furthermore, this type of kernel may not work well for large feature spaces.\footnote{Appendix C of \citet{hsu2003practical} examines this issue in greater detail.}
Although they do not perform as well as linear SVMs for text classification, they can achieve very competitive results on problems with fewer features.

\subsubsection{Logistic Regression}
Logistic regression is a type of linear regression model where the dependent variables are categorical.
Supported by strong theoretical underpinnings as well as practical outcomes, maximum likelihood logistic regression has become a widely used machine learning algorithm. 
Although high-dimensional input poses a challenge for these models \citep{genkin2007large}, this issue can be addressed to some degree using regularization methods \citep{zhu2004classification}.
This algorithm is inherently multi-class, meaning that OVA and OVO approaches are not required.
The logistic regression classifier is also probabilistic and provides continuous probability estimates for each class label.

\subsubsection{Perceptron}
The Perceptron \cite{rosenblatt1958perceptron} is another linear learning algorithm that has been successful
The algorithm learns a weight vector and a bias term which shifts the decision boundary from the origin. However, the algorithm will not converge if the data is not linearly separable.
It supports online learning and
each training instance is processed and weights updated according to a defined learning rate.
Perceptrons have been successfully used for POS tagging \cite{W02-1001} and parsing \cite{P04-1015}.

\subsubsection{Ridge Regression}
Classification using ridge regression is an approach based on a regression model that uses a linear least squares loss function \cite{zhang2001text}.
The OVA approach is used for multiclass classification.
Given that it is a linear model, it can work well for high-dimensional problems as they are often linearly separable.

\subsubsection{Decision Trees}
One of the oldest and most commonly used supervised learning method, decision trees are a non-parametric method that attempt to learn a set of hierarchical decision rules based on the input features \citep{quinlan1993}.
They are inherently multiclass learners and require little data preprocessing.
However, the trees can be unstable and may not generalize well beyond the training data.

\subsubsection{Linear Discriminant Analysis}
A classic learning algorithm, Linear Discriminant Analysis (LDA, not to be confused with Latent Dirichlet Allocation) is a method based on a linear decision boundary \cite{fisher1936use}. It has been widely and successfully used for classification \cite{liu2002gabor}.
LDA, a generative classification method,  fits a conditional probability density function to each class 
and works under the assumption of homoscedasticity, \ie all classes have the same covariance.
It is non-parametric and inherently multiclass.

\subsubsection{Quadratic Discriminant Analysis}
Quadratic Discriminant Analysis (QDA) is similar to LDA, except that it uses a quadratic decision surface \cite{hastie:2009}.
Unlike LDA, however, it makes no assumption about equal class covariances, allowing them to be class-specific.

\subsubsection{$k$-nearest Neighbors}
A popular neighbor-based algorithm, $k$-nearest Neighbors (k-NN) is an instance-based classifier that does not build a statistical model \citep{cover1967nearest}.
Training data are stored and test instances are labelled through a majority vote of the labels of the $k$ nearest instances.
The $k$ parameter must be defined. This value is usually data-dependent and chosen experimentally.

\subsubsection{Nearest Centroid}
The Nearest Centroid (NC) classification algorithm computes the centroid (\ie mean) vector for each class \citep{tibshirani2002diagnosis}. Test instances are assigned to the class with the closest centroid.
It is non-parametric but can perform poorly when classes have different variances for each feature.

\subsection{Classifier Output Representation}
\label{sec:clfoutput}

As we will describe in \S\ref{sec:models}, our meta-classifiers are trained on the outputs generated by individual classifiers.
This output generally falls into two categories: discrete labels and continuous values. In this section we briefly describe these and how they are used for further classification.

\subsubsection{Discrete Label Values}

The most elementary approach is to use the discrete class labels produced by the classifiers.
In the case of multi-class classification this output is a single discrete value representing the hypothesis formed by the classifier and is available from virtually all learning algorithms. The number of possible values is the same as the number of possible classes in the dataset, $K$.

To use this categorical value as a classification feature, the outputs from any classifier
must be represented as a feature vector.
A common approach here to represent the values using one-hot encoding. This encoding creates a 1-of-$K$ vector: this is a vector with $K$ elements where one element will always be set to $1$ while the rest are $0$.
This approach enables categorical data to be represented as continuous input, which is the input format expected by most learning algorithms.

\subsubsection{Continuous Output}

Many classification algorithms can also produce continuous output associated with each class.
This output can represent the confidence for each class label.
Probabilistic classifiers, such as Logistic Regression, can provide confidence estimates for each of the possible $K$ class labels.
Margin-based classifiers, like Support Vector Machines, can provide the signed distance to the separating hyperplane.\footnote{There exist additional methods to map these distances to probabilities \citep{platt1999probabilistic}.}

Where available, this output information can also be used to form a vector for classification. For each input, this would result in a $K$-element vector where each element is the continuous output associated with a class label.
For confidence estimates all elements in the vector would sum to $1$.

The confidence levels for each label can provide useful information; by considering the values for the other labels we may be able to make better predictions.
They can also help prevent voting tie issues that can occur when only using the discrete class labels.

\section{Features}
\label{sec:features}

This study utilizes a standard set of NLI features widely used in previous work.
This study focuses on comparing classification methodology for NLI and we do not confound this objective by introducing new features.
Different feature types are extracted from each of our three datasets, as shown in Table~\ref{tab:feats}.

\begin{table}
\caption{An overview of the features available for each dataset.}
\center
\begin{tabular}{rccc}
\hline\hline
\textbf{Feature}	& \textbf{English} &  \textbf{Chinese}	& \textbf{Norwegian} \\ 
\hline
Word/Lemma \ngs 	&	X	&	& \\
Character \ngs		&	X	&	& \\
Function word unigrams	&	X	&	X	&	X \\
Function word bigrams	&	X	&	X	&	X \\
Part-of-Speech (POS) \ngs	&	X	&	X	&	X \\
Dependencies	&	X	&	X	&	\\
CFG Rules		&	X	&	X	&	\\
Adaptor Grammars	&	X	&	&	\\
TSG Fragments	&	X	&	&	\\
\hline\hline
\label{tab:feats}
\end{tabular}
\end{table}

The feature types for each dataset were chosen based on properties of the dataset and the availability of NLP resources for the L2.

For stylistic classification tasks like NLI, these content-based features can only be used if the training data is balanced for topic. Otherwise, topic balance will greatly impact the results and artificially inflate accuracy \cite{malmasi:2015:mnli}.
Accordingly, we only use these features for our experiments with \tfl, which is balanced for topic.
The NLP tools used to extract adaptor grammar and TSG fragment features are only available for English and thus limited to \tfl.
The paucity of NLP tools for Norwegian and the lack of topic balance in the data also limited our features to those manually annotated in the dataset.

This imbalance of feature types across the corpora is not an issue as we are comparing the performance of our models \textit{within} each dataset and not between the L2s.

The remainder of this section describes the feature types listed in Table~\ref{tab:feats}.

\subsection{Word, Lemma and Character n-grams}

We extract commonly used surface features from the texts. These include word unigrams and bigrams, lemma unigrams and bigrams and character uni/bi/trigrams.

\subsection{Function Words}

In contrast to content words, function words do not have any thematic meaning themselves, but rather can be seen as indicating the grammatical relations between other words.
In a sense, they are the syntactic glue that hold much of the content words together and their role in assigning syntax to sentences is linguistically well-defined. 
They generally belong to a language's set of closed-class words and
embody relations more than propositional content.  Examples include
articles, determiners, conjunctions and auxiliary verbs.

Function words are considered to be highly context- and topic-independent but other open-class words can also exhibit such such properties.  In
practical applications, such as Information Retrieval, such words are
often removed as they are not informative and stoplists for different
languages have been developed for this purpose.  These lists contain
`stop words' and formulaic discourse expressions such as
\textit{above-mentioned} or \textit{on the other hand}.

Function words' topic independence has led them to be widely used in studies of
authorship attribution \citep{Mosteller:1964} as well as
NLI\footnote{For example, the largest list used by
  \citet{Wong:2009} was a stopword list from Information Retrieval;
  given the size of their list, this was presumably also the case for 
\citet{Koppel:2005}, although the source there was not given.}
 and
they have been established to be informative for these tasks. Much
like Information Retrieval, the function word lists used in these
tasks are also often augmented with stoplists and this is also the
approach that we take.

Such lists generally contain anywhere from 50 to several hundred words, depending on the granularity of the list and also the language in question.
In this work, the English word list was obtained from the Onix Text Retrieval Toolkit.\footnote{\url{http://www.lextek.com/manuals/onix/stopwords1.html}}
For Norwegian we used a list of $176$ function words obtained from the distribution of the Apache Lucene search engine software.\footnote{\url{https://github.com/apache/lucene-solr}}
This list includes stop words for the Bokm\r{a}l variant of the language and contains entries such as \textit{hvis} (whose), \textit{ikke} (not), \textit{jeg} (I), \textit{s\r{a}} (so) and \textit{hj\r{a}} (at). We also make this list available on our website.\footnote{\url{http://web.science.mq.edu.au/~smalmasi/data/norwegian-funcwords.txt}}
For Chinese, we utilize the function word list described in \citet{malmasi:2014:cnli}.

In addition to single function words, we also extract function word bigrams, as described by \citet{malmasi-wong-dras:2013:BEA8}.
Function word bigrams are a type of word \textit{n}-gram where content words
are skipped: they are thus a specific subtype of the skip-grams discussed by \citet{guthrie-etal:2006:LREC}.
For example, the sentence
``\textit{We should all start taking the bus}" would be reduced to
``\textit{we should all the}", from which we would extract the $n$-grams.

\subsection{Part-of-Speech n-grams}

Parts of Speech (POS) are linguistic categories (or word classes) assigned to words that signify their syntactic role. Basic categories include verbs, nouns and adjectives but these can be expanded to include additional morpho-syntactic information. The assignment of such categories to words in a text adds a level of linguistic abstraction.

We extract POS $n$-grams of order $1$--$3$, which have been shown to be useful for NLI \cite{malmasi-wong-dras:2013:BEA8}.
These \ngs capture small and very local syntactic patterns of language production and were used as classification features.
Previous work and our experiments showed that sequences of size $4$ or greater achieve lower accuracy, possibly due to data sparsity, so we do not include them.

For English and Chinese, the Stanford
CoreNLP\footnote{\url{http://nlp.stanford.edu/software/corenlp.shtml}}
  suite of NLP tools \citep{CORENLP} and the provided models were used
  to tokenize, POS tag and parse the unsegmented corpus texts.
We did not use any NLP tools for Norwegian as the corpus we use is already annotated with POS tags.

Additionally, we extract a second set of POS \ngs for the \tfl data using the CLAWS dataset, which has been shown to perform well for NLI \cite{malmasi-wong-dras:2013:BEA8}.

\subsection{Adaptor grammar collocations} 
For the \tfl data, we utilize an
adaptor grammar to discover arbitrary length $n$-gram
collocations. 
We explore both the pure
part-of-speech (POS) $n$-grams as well as the more promising
mixtures of POS and function words.
We derive
two adaptor grammars where each is associated with a different set of
vocabulary: either pure POS or the mixture of POS and function
words. We use the grammar proposed by \cite{Johnson:2010:ACL} for
capturing topical collocations:

\begin{center}
\scalebox{0.8}{
\begin{tabular}{ll}
	$Sentence \rightarrow Doc_j$ & $j \in 1, \ldots, m$ \\ $Doc_j
  \rightarrow \_j$ & $j \in 1, \ldots, m$ \\ $Doc_j \rightarrow
  Doc_j \hspace{1mm} Topic_i$ & $i \in 1, \ldots, t;$ \\ & $j \in 1,
  \ldots, m$ \\ $\underline{Topic_i} \rightarrow Words$ & $i \in 1,
  \ldots, t$ \\ $Words \rightarrow Word$ & \\ $Words \rightarrow
  Words \hspace{1mm} Word$ & \\ $Word \rightarrow w$ & $w \in
  V_{pos};$ \\ & $w \in V_{pos+fw}$\\
\end{tabular}
}
\end{center}

$V_{pos}$ contains 119 distinct POS tags based on the Brown tagset and $V_{pos+fw}$ is extended with 398 function words.
The number of topics $t$ is set to $50$.
The inference algorithm for the adaptor grammars are based on the Markov Chain Monte Carlo technique made available by \citet{Johnson:2010:ACL}.\footnote{\url{http://web.science.mq.edu.au/~mjohnson/Software.htm}}

\subsection{Stanford dependencies}
For English and Chinese we use Stanford dependencies as a syntactic feature:
 for each text we extract all the basic dependencies returned by the Stanford Parser \citep{deMarneffe-Maccartney-Manning:2006:LREC}. We then generate all the variations for each of the dependencies (grammatical relations) by substituting each lemma with its corresponding POS tag. For instance, a grammatical relation of \texttt{det(knowledge, the)} yields the following variations: \texttt{det(NN, the)}, \texttt{det(knowledge, DT)},  and \texttt{det(NN, DT)}.

\subsection{CFG Rules}
\label{sec:nlifeat-pr}

\begin{figure}
\centering
\includegraphics[width=0.80\textwidth]{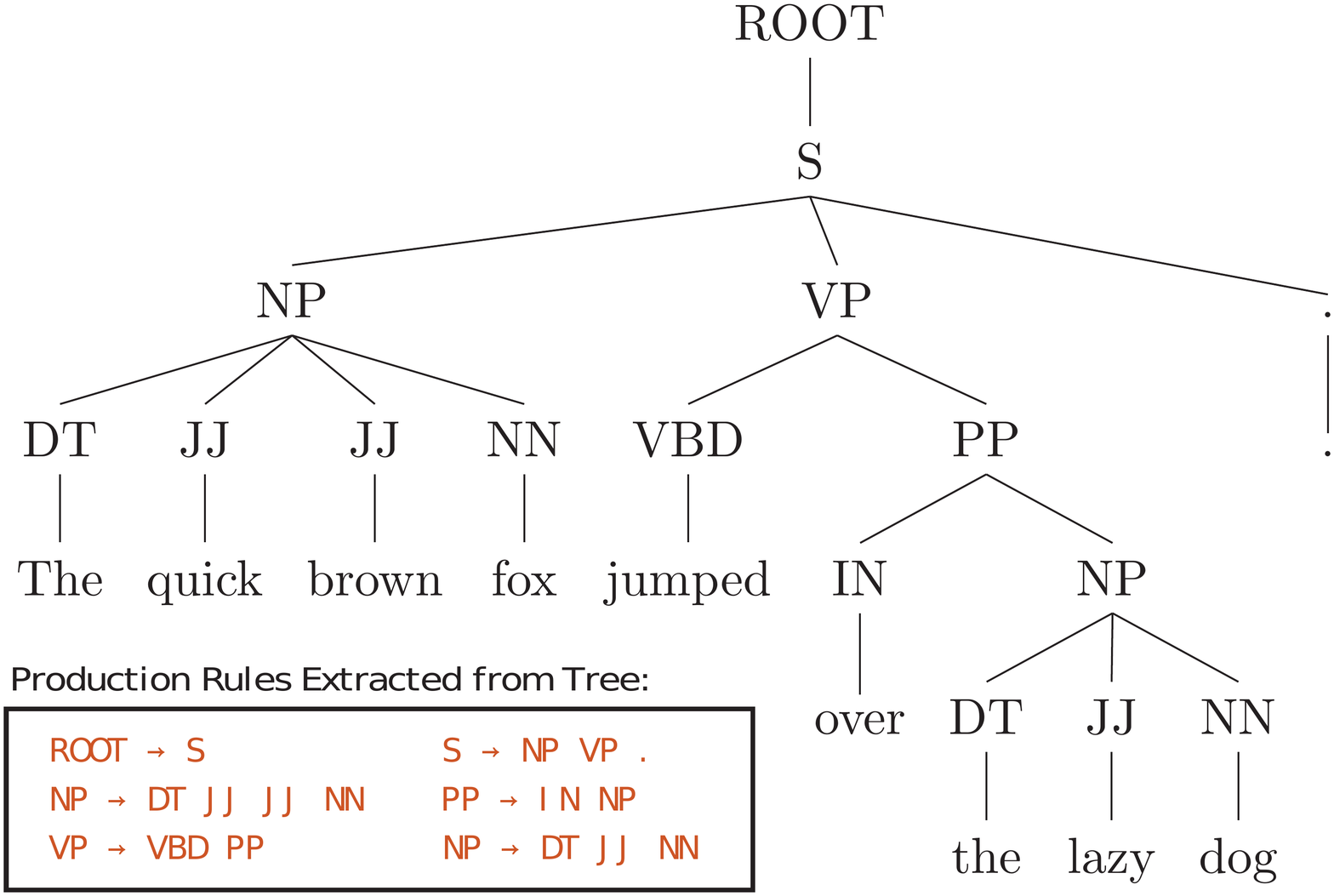}
\caption[A parse tree and its extracted production rules.]{A constituent parse tree for an example sentence along with the context-free grammar production rules which can be extracted from it.}
\label{fig:prodrules1}
\end{figure}

Also known as Phrase Structure Rules or Production Rules, these are the rules used to generate constituent parts of sentences, such as noun phrases.
One way to obtain these is by first generating constituent parses for all sentences.
The production rules, excluding lexicalizations, are then extracted.
Figure \ref{fig:prodrules1} illustrates this with an example tree and its rules. 

These context-free phrase structure rules capture the overall structure of grammatical constructions and global syntactic patterns. They can also encode highly idiosyncratic constructions that are particular to some L1 group. They have been found to be useful for NLI \citep{wong-dras:2011:EMNLP}.
We use the Stanford parser to extract these features for both English and Chinese.

\subsection{Tree Substitution Grammar Fragments}
\label{sec:nlifeat-tsg}

Tree Substitution Grammar (TSG) fragments have been proposed by
\citet{swanson-charniak:2012:ACL2012short} as yet another type of
syntactic feature for NLI or other syntactically motivated text classification tasks.
They demonstrated that this feature type can achieve high classification accuracy.

\begin{figure}
\centering
\includegraphics[width=0.65\textwidth]{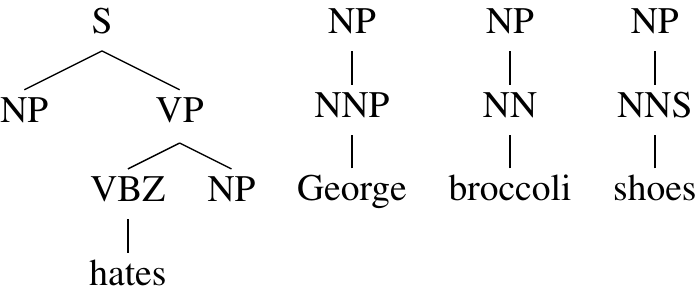}
\caption[A example of a Tree Substitution Grammar.]{Fragments from a Tree Substitution Grammar capable of deriving the sentences ``George hates broccoli" and ``George hates shoes". Reproduced from \citet{swanson-charniak:2012:ACL2012short}.}
\label{fig:tsg}
\end{figure}

TSGs are a generalization of context-free grammars that allow
non-terminals to rewrite as fragments which can have an arbitrary size \citep{post2013bayesian}, instead of being limited to a depth of one.
A TSG \textit{fragment} or \textit{elementary tree} refers to these rules.
Figure~\ref{fig:tsg} shows several example fragments from a Tree Substitution Grammar capable of deriving the sentences ``George hates broccoli" and ``George hates shoes".
We only extract TSG fragments for the \tfl data as they include lexical terminal nodes.

\section{Classification Models}
\label{sec:models}

We conduct a set of three experiments, each based on different ensemble structures which we describe in this section.
The first model is based on a traditional parallel ensemble structure while the second model examines meta-classification using classifier stacking.
The third and final model is a hybrid approach, building an ensemble of meta-classifiers.

\subsection{Ensemble Classifiers}
\label{sec:ensemble-intro}

The most common ensemble structure, as described earlier in \S\ref{sec:ensemble-relwork}, relies on a set of base classifiers whose decisions are combined using some predefined method. This is the approach for our first model.

\begin{figure}
\centering
\includegraphics[trim=0 0 0 0,clip,width=1\textwidth]{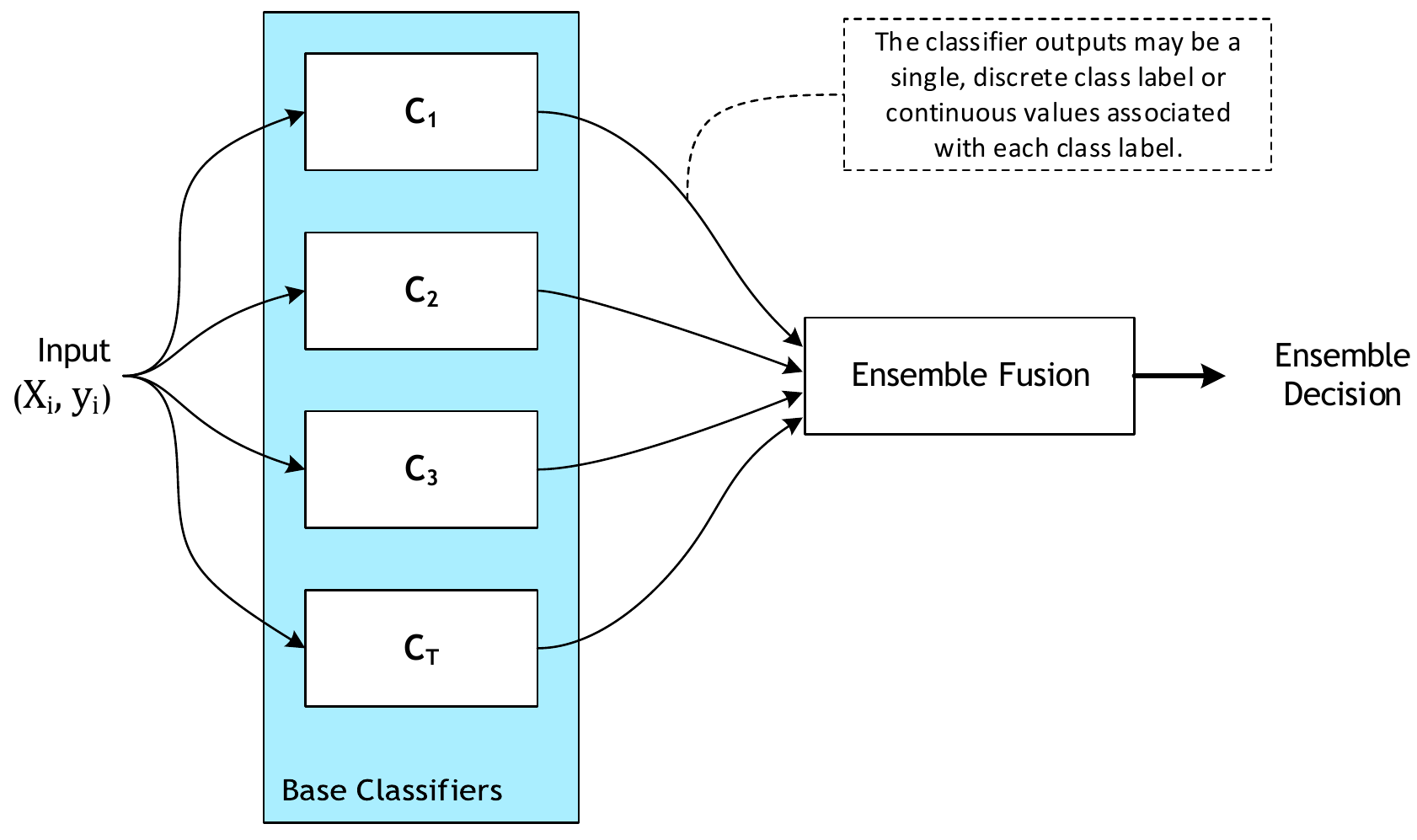}
\caption[Parallel classifier ensemble architecture]{An example of a  parallel ensemble classifier architecture where $T$ independent classifiers provide predictions which are then fused using a rule-based ensemble combination method. The class labels for the input, $y_i$, are only available during training or cross-validation.}
\label{fig:ensemble-architecture}
\end{figure}

Such systems often use a parallel architecture, as illustrated in Figure \ref{fig:ensemble-architecture}, where the classifiers are run independently and their outputs are aggregated using a fusion method.
The first part of creating an ensemble is generating the individual classifiers.
Various methods for creating these ensemble elements have been proposed.
These involve using different algorithms, parameters or feature types; applying different preprocessing or feature scaling methods; and varying (\eg  distorting or resampling) the training data.

For example, \textit{Bagging} (bootstrap aggregating) is a commonly used method for ensemble generation \citep{Breiman96baggingpredictors} that can create multiple base classifiers.
It works by creating multiple bootstrap training sets from the original training data and a separate classifier is trained from each set.
The generated classifiers are said to be diverse because each training set is created by sampling with replacement and contains a random subset of the original data.
\textit{Boosting} (\eg with the AdaBoost algorithm) is another method where the base models are created with different weight distributions over the training data with the aim of assigning higher weights to training instances that are misclassified \citep{freund1996experiments}.

In our first model we follow the same approach as previous NLI research and train each classifier on a different feature type.\footnote{This can also be achieved through training each classifier on a subspace of the entire feature set that includes all types.}
However, our other models make use of the boosting and bagging techniques, which will be discussed in later sections.
For reasons given in \S\ref{sec:svm}, we use linear SVMs for these base classifiers.

The second part of ensemble design is choosing a combination or fusion rule to aggregate the outputs from the various learners, this is discussed in the next section.
Most research to date has not compared different types of such combiners and we aim to evaluate a number of different strategies.

\subsubsection{Ensemble Fusion Methods}
\label{sec:ensemble-combiners}

Once it has been decided how the set of base classifiers will be generated,
selecting the classifier combination method is the next fundamental design question in ensemble construction.

The answer to this question depends on what output is available from the individual classifiers.
The two different output types were discussed earlier in \S\ref{sec:clfoutput}.
Some combination methods are designed to work with class labels, assuming that each learner outputs a single class label prediction for each data point.
Other methods are designed to work with class-based continuous output, requiring that for each instance every classifier provides a measure of confidence\footnote{\eg an estimate of the posterior probability for the class label.} for each class label. These outputs may correspond to probabilities for each class and consequently  sum to $1$ over all the classes.
If an algorithm can provide both types of output, then all the methods can be tested. This is the case for the classifiers we will work with, as they are all SVMs.

These methods are usually based on some predefined rule or logic and cannot be trained. This can be considered an advantage, allowing them to be implemented and used without additional training of domain-specific combination models. On the other hand, they may not be able to exploit domain-specific trends and patterns in the input data.

Although a number of different fusion methods have been proposed and tested, there is no single dominant method \citep{polikar2006ensemble}.
The performance of these methods is influenced by the nature of the problem and available training data, the size of the ensemble, the base classifiers used and the diversity between their outputs.
This is an important motivation for comparatively assessing these methods on NLI data.

The selection of this method is often done empirically.
Many researchers have compared and contrasted the performance of combiners on different problems, and most of these studies -- both empirical and theoretical -- do not reach a definitive conclusion \citep[p 178]{kuncheva2014combining}.

In the same spirit, we experiment with several classifier fusion methods which have been widely applied and discussed in the machine learning literature.
Our selected methods are described below; 
a variety of other methods exist and the interested reader can refer to the thorough exposition by \cite{polikar2006ensemble}.

\subsubsection{Plurality voting}
\label{sec:ens-vote}

Each classifier votes for a single class label. The votes are tallied and the label with the highest number of votes wins.\footnote{This differs with a \textit{majority} voting combiner where a label must obtain over $50\%$ of the votes to win. However, the names are sometimes used interchangeably.} Ties are broken arbitrarily.
This method is simple and does not have any parameters to tune.
An extensive analysis of the method and its theoretical underpinnings can be found in \citet[p. 112]{kuncheva2004combining}.

\subsubsection{Mean Probability Rule}
\label{sec:ens-mean}

\begin{figure}
\centering
\includegraphics[width=0.7\textwidth]{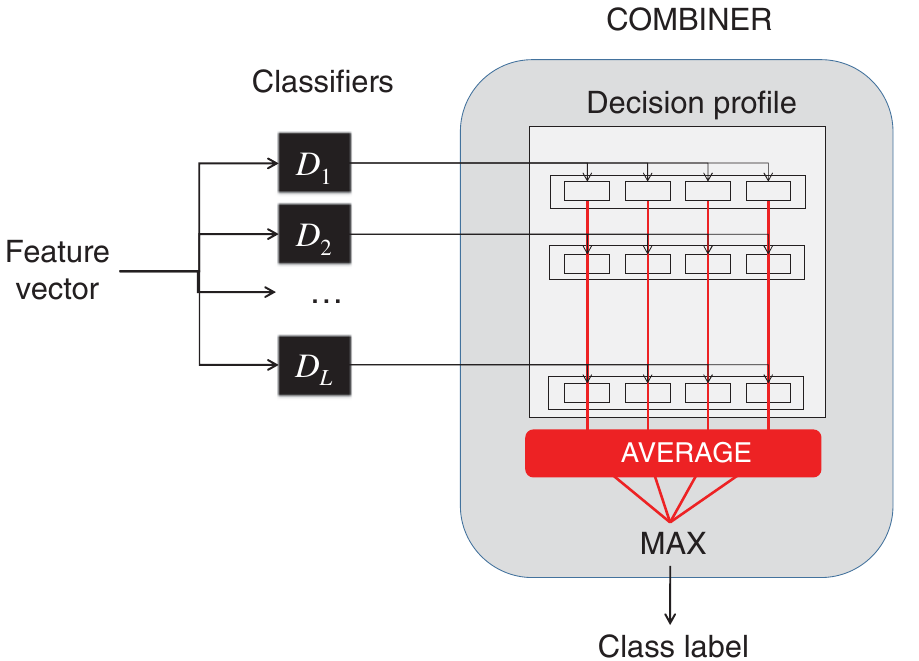}
\caption[The mean probability ensemble combiner.]{An example of a mean probability combiner. The feature vector for a sample is input to $L$ different classifiers, each of which output a vector of confidence probabilities for each possible class label. These vectors are combined to form the decision profile for the instance which is used to calculate the average support given to each label. The label with the maximum support is then chosen as the prediction. Image reproduced from \cite[Fig. 5.5]{kuncheva2014combining}.}
\label{fig:ensemble-average-combiner}
\end{figure}

The probability estimates for each class, provided by each individual classifier, are summed and the class label with the highest average probability is the winner. This is illustrated in Figure~\ref{fig:ensemble-average-combiner}. This is equivalent to the probability sum combiner which does not require calculating the average for each class.
An important aspect of using probability outputs in this way is that a classifier's support for the true class label is taken into account, even when it is not the predicted label (\eg it could have the second highest probability).
This method has been shown to work well on a wide range of problems and, in general, it is considered to be simple, intuitive, stable \citep[p. 155]{kuncheva2014combining} and resilient to estimation errors \citep{kittler1998combining}, making it one of the more robust combiners discussed in the literature.

\subsubsection{Median Probability Rule}
\label{sec:ens-median}

Given that the mean probability used in the above rule is sensitive to outliers, an alternative is to use the median as a more robust estimate of the mean \citep{kittler1998combining}.
Under this rule, each class label's estimates are sorted and the median value is selected as the final score for that label. The label with the highest median value is picked as the winner.
As with the mean combiner, this method measures the central tendency of support for each label as a means of reaching a consensus decision.

\subsubsection{Product Rule}
For each class label, all of the probability estimates are multiplied together to create the label's final estimate \citep[p. 37]{polikar2006ensemble}. The label with the highest estimate is selected.
This rule can theoretically provide the best overall estimate of posterior probability for a label, assuming that the individual estimates are accurate.
A trade-off here is that this method is very sensitive to low probabilities: a single low score for a label from any classifier will essentially eliminate that class label.

\subsubsection{Highest Confidence} In this simple method, the class label that receives the vote with the largest degree of confidence is selected as the final prediction \citep[p. 150]{kuncheva2014combining}.
In contrast to the previous methods, this combiner disregards the consensus opinion and instead picks the prediction of the expert with the highest degree of confidence.

\subsubsection{Borda Count}
This method works by using each classifier's confidence estimates to create a ranked list of the class labels in order of preference, with the predicted label at rank $1$.
The winning label is then selected using the Borda count\footnote{This method is generally attributed to Jean-Charles de Borda ($1733$--$1799$), but evidence suggests that it was also proposed by Ramon Llull ($1232$--$1315$).} algorithm
\citep{ho1994decision}.
The algorithm works by assigning points to labels based on their ranks.
If there are $N$ different labels, then each classifier's preferences are assigned points as follows: the top-ranked label receives $N$ points, the second place label receives $N-1$ points, third place receives $N-2$ points and so on with the last preference receiving a single point.
These points are then tallied to select the winner with the highest score.

The most obvious advantage of this method is that it takes into account all of each classifier's preferences, making it possible for a label to win even if another label received the majority of the first preference votes.

\subsubsection{Oracle Combiners}
\label{sec:oracle}

Our final set of combiners are designed to assist with assessing the potential performance upper bound that could be achieved by a system, given a set of classifiers. As such, they are primarily used for evaluation in the same manner that baselines help determine the lower bounds for performance.
These combiners cannot be used to make predictions on unlabelled data.

One possible approach to estimating an upper-bound for classification accuracy, and one that we employ here, is the use of an ``Oracle" combiner.
This method has previously been used to analyze the limits of majority vote classifier combination \citep{kuncheva2001decision}.
An oracle is a type of multiple classifier fusion method that can be used to combine the results of an ensemble of classifiers which are all used to classify a dataset.

The oracle will assign the correct class label for an instance if at least one of the constituent classifiers in the system produces the correct label for that data point.
Some example oracle results for an ensemble of three classifiers are shown in Table~\ref{tab:oracle-example}.
The probability of correct classification of a data point by the oracle is:
$$P_{\text{Oracle}} = 1 - P(\text{All Classifiers Incorrect})$$

Oracles are usually used in comparative experiments and to gauge the performance and diversity of the classifiers chosen for an ensemble \citep{kuncheva2002theoretical,kuncheva2003limits}.
They can help us quantify the \textit{potential} upper limit of an ensemble's performance on the given data and how this performance varies with different ensemble configurations and combinations.

\begin{table}[t]
\caption{Example oracle results for an ensemble of three classifiers.}
\label{tab:oracle-example}
\centering
\scalebox{0.85}{
\begin{tabular}{lccccl}
\hline\hline
& & \multicolumn{3}{c}{Classifier Output} & \\
\textbf{Instance} & \textbf{True Label} & \textbf{$C_1$} & \textbf{$C_{2}$} & \textbf{$C_{3}$} & \textbf{Oracle} \\
\hline
\tt{18354.txt} & \tt{ARA} & \tt{TUR} & \tt{ARA} & \tt{ARA} & Correct \\
\tt{15398.txt} & \tt{CHI} & \tt{JPN} & \tt{JPN} & \tt{KOR} & Incorrect \\
\tt{22754.txt} & \tt{HIN} & \tt{GER} & \tt{TEL} & \tt{HIN} & Correct \\
\tt{10459.txt} & \tt{SPA} & \tt{SPA} & \tt{SPA} & \tt{SPA} & Correct \\
\tt{11567.txt} & \tt{ITA} & \tt{FRE} & \tt{GER} & \tt{SPA} & Incorrect \\
\hline\hline
\end{tabular}
}
\end{table}

To account for the possibility that a classifier may predict the correct label essentially by chance (with a probability determined by the random baseline) and thus exaggerate the oracle score, we also use an Accuracy@$N$ combiner.
This method is inspired by the ``Precision at $k$" metric from Information Retrieval \citep{manning:2008:eval} which measures precision at fixed low levels of results (\textit{e.g.} the top $10$ results).
Here, it is an extension of the Plurality vote combiner where instead of selecting the label with the highest votes, the labels are ranked by their vote counts and an instance is correctly classified if the true label is in the top $N$ ranked candidates.\footnote{In case of ties, we choose randomly from the labels with the same number of votes.}
Another way to view it is as
a more restricted version of the Oracle combiner that is limited to the top $N$ ranked candidates in order to minimize the influence of a single classifier having chosen the correct label by chance.
In this study we experiment with $N = 2$ and $3$. We also note that setting $N = 1$ is equivalent to the Plurality voting method and setting $N$ to the number of class labels is equivalent to the Oracle combiner.

\subsection{Meta-Classifiers (Stacked Generalization)}
\label{sec:ens-sg}

While the combination methods in our first model are not trainable, other more sophisticated ensemble methods that rely on meta-learning employ a stacked architecture where the output from a first layer of classifiers is fed into a second level meta-classifier and so on.
For our second model we expand our methodology to such a meta-classifier, also referred to as \textit{stacked generalization} or \textit{classifier stacking} \cite{wolpert1992stacked}.
This methodology has not been tested for NLI thus far.

\begin{figure}[h]
\centering
\includegraphics[trim=0 0 0 0,clip,width=1\textwidth]{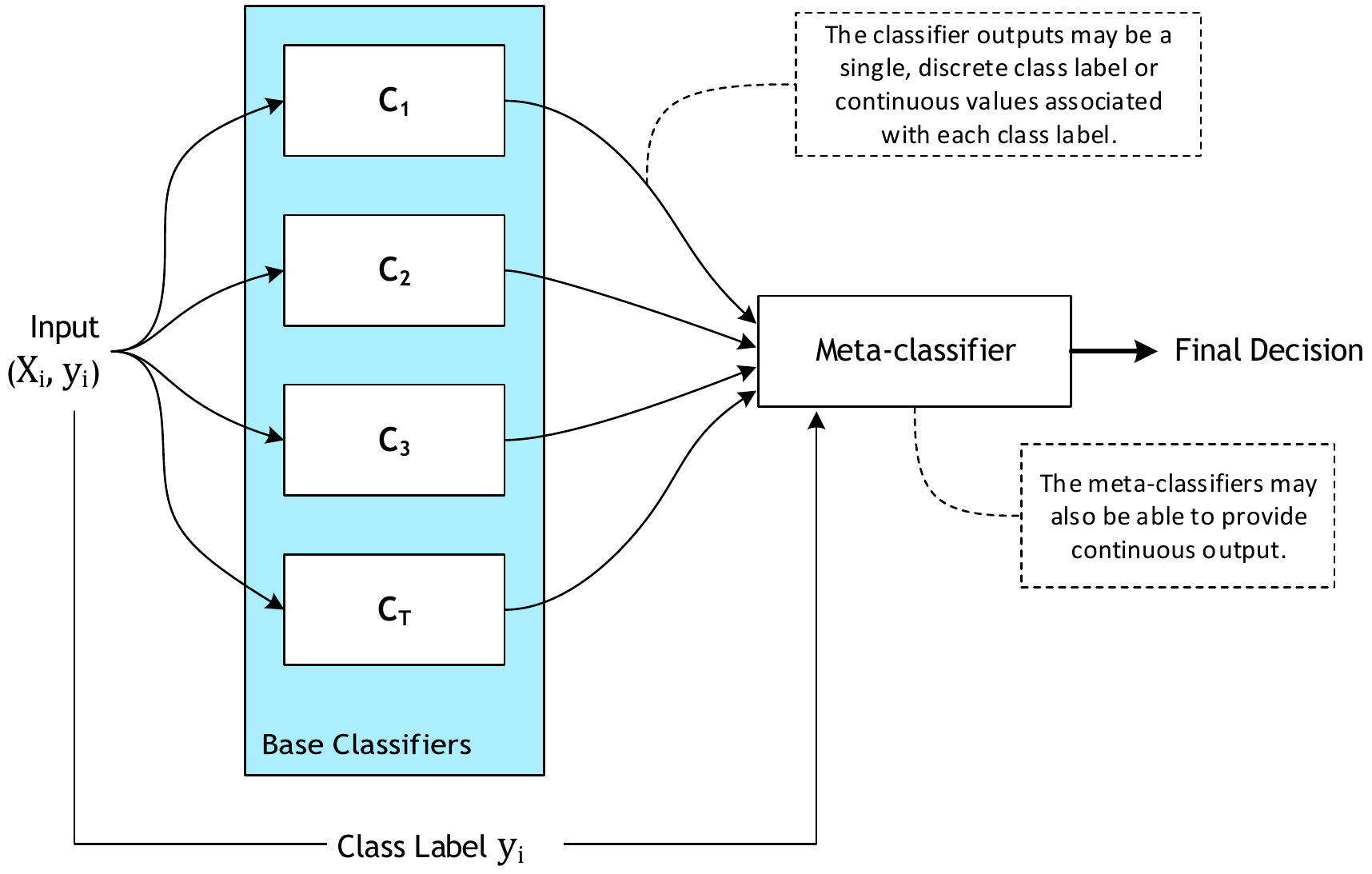}
\caption{A example architecture for a meta-classifier. The outputs generated by the set of $T$ base classifiers, along with the labels of the training data, are used to train a meta-learner that can predict the final decision for an input. This meta-classifier attempts to learn from the collective knowledge represented by the ensemble of base classifiers and may be able to learn and exploit regularities in their output. The class labels for the input are only available during training or cross-validation.}
\label{fig:metaclassifier}
\end{figure}

A meta-classifier architecture is composed of an ensemble of base classifiers, just as in our first model.
A key difference is that instead of employing a rule-based fusion method, the individual classifier outputs, along with the training labels, are used to train a second-level meta-classifier.
This second meta-learner serves to predict the final decision for an input, given the decisions of the base classifiers.
This setup is illustrated in Figure~\ref{fig:metaclassifier}.

This meta-classifier attempts to learn from the collective knowledge represented by the ensemble of local classifiers and may be able to learn and exploit patterns and regularities in their output \cite[\S3.6]{polikar2006ensemble}.
For example, it may be the case that a certain ensemble element is more accurate at classifying instances of a certain class or there may be interactions or correlations between the outputs of certain elements that could help improve results over a simple fusion method.
So the meta-classifier may learn how the base classifiers commit errors and attempt to correct their biases.

Just as there are different fusion methods for ensemble combination, different learning algorithms can be used for the meta-classifier element. 
In this study we experiment with all of the learning algorithms listed earlier in \S\ref{sec:learners}.
Additionally, we also test each learner using both the continuous and discrete output data in order to comparatively assess their performance.
This approach allows us to evaluate whether one method performs better, and if certain algorithms are better suited for some specific input formats.

Similar to the base learners, the meta-classifier can generate both continuous and discrete output. 
In this model we take the discrete label output and use it as the final decision to be used in evaluating the model.

For training, the input for the meta-classifier can be obtained from the outputs of the base classifiers under cross-validation. That is to say, the classifier outputs from each test fold are paired with the original gold-standard label and this is used to train the meta-classifier.

\subsection{Meta-Classifier Ensembles}
\label{sec:ens-mcens}

The two models described thus far have relied on multiple classifier combination and meta-learning. While they both have their advantages, would it be possible to combine both approaches?

\begin{figure}
\center
\includegraphics[trim=0 0 0 0,clip,width=1\textwidth]{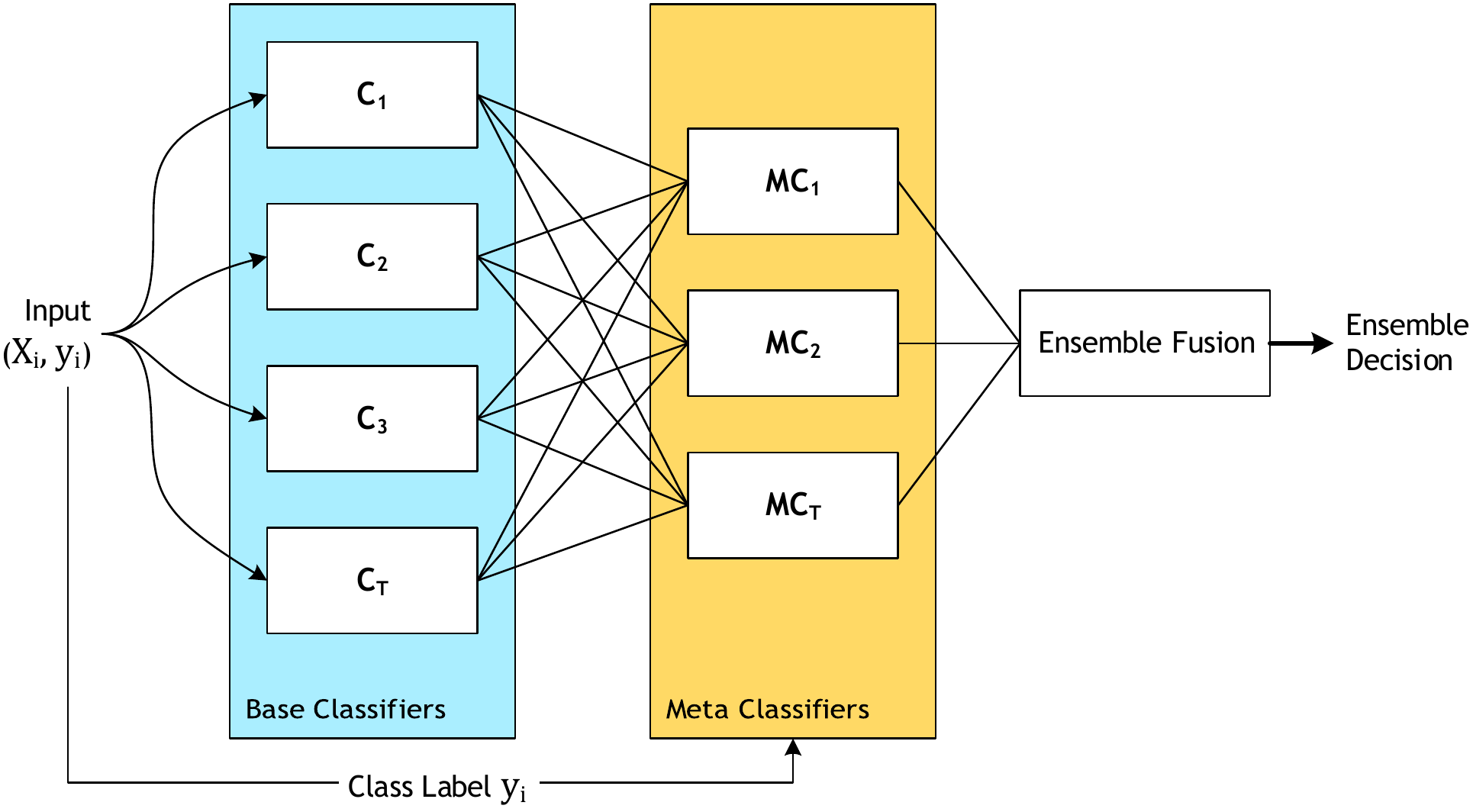}
\caption{An illustration of our meta-classifier ensemble. The outputs from the ensemble of base classifiers is used to train an ensemble of meta-classifiers, the output of which is processed using a decision fusion method. This is a hybrid approach that attempts to combine both ensemble fusion and classifier stacking.}
\label{fig:metac-ensemble}
\end{figure}

To this end, our final model is a hybrid of the previous two approaches that attempts to answer this question.
Results from the set of base classifiers are provided to an ensemble of meta-classifiers, instead of a single one. The outputs from the meta-classifiers are then combined using a fusion method to reach a final verdict.
The layout for this model is illustrated in Figure~\ref{fig:metac-ensemble}.
This approach, while adding substantial complexity to the model, could potentially combine the benefits of stacking and ensemble combination.

Additionally, this approach also requires a method for generating the meta-classifier ensemble itself. While the first level ensemble is generated using a different feature type per classifier, that method cannot be applied here since we are using classifier outputs.
For that purpose we experiment with boosting and bagging (as described in \S\ref{sec:ensemble-intro}).
These methods have been widely used for creating decision tree ensembles \citep{geurts2006extremely};
we will experiment with random forests, extra trees and the AdaBoost algorithm.
We will also experiment with bagging, which can be applied to any learner that can be applied for meta-classification, such as SVMs.

\section{Experiments and Results}
\label{sec:results}

We divide our experiments into two parts: comprehensive experiments on the English \tfl data (\S\ref{sec:exp1}), followed by comparative experiments on Chinese and Norwegian data (\S\ref{sec:exp2}).

\subsection{Experiments on TOEFL11}
\label{sec:exp1}

Given that it is the largest and most widely used corpus, we evaluate all of our models on \tfl, before comparing their performances on the other datasets.

Results are compared against a random baseline and the oracle combiners.
We also compare these results against the winning system from the 2013 NLI shared task \cite{jarvis-bestgen-pepper:2013:BEA8} and two systems by \cite{bykh:2014} and \cite{ionescu:2014:nli} which presented state-of-the-art results following the task. They were all previously described in \S\ref{sec:relwork}.

\begin{figure}
\centering
\includegraphics[trim=5 10 15 55,clip,width=0.6\textwidth]{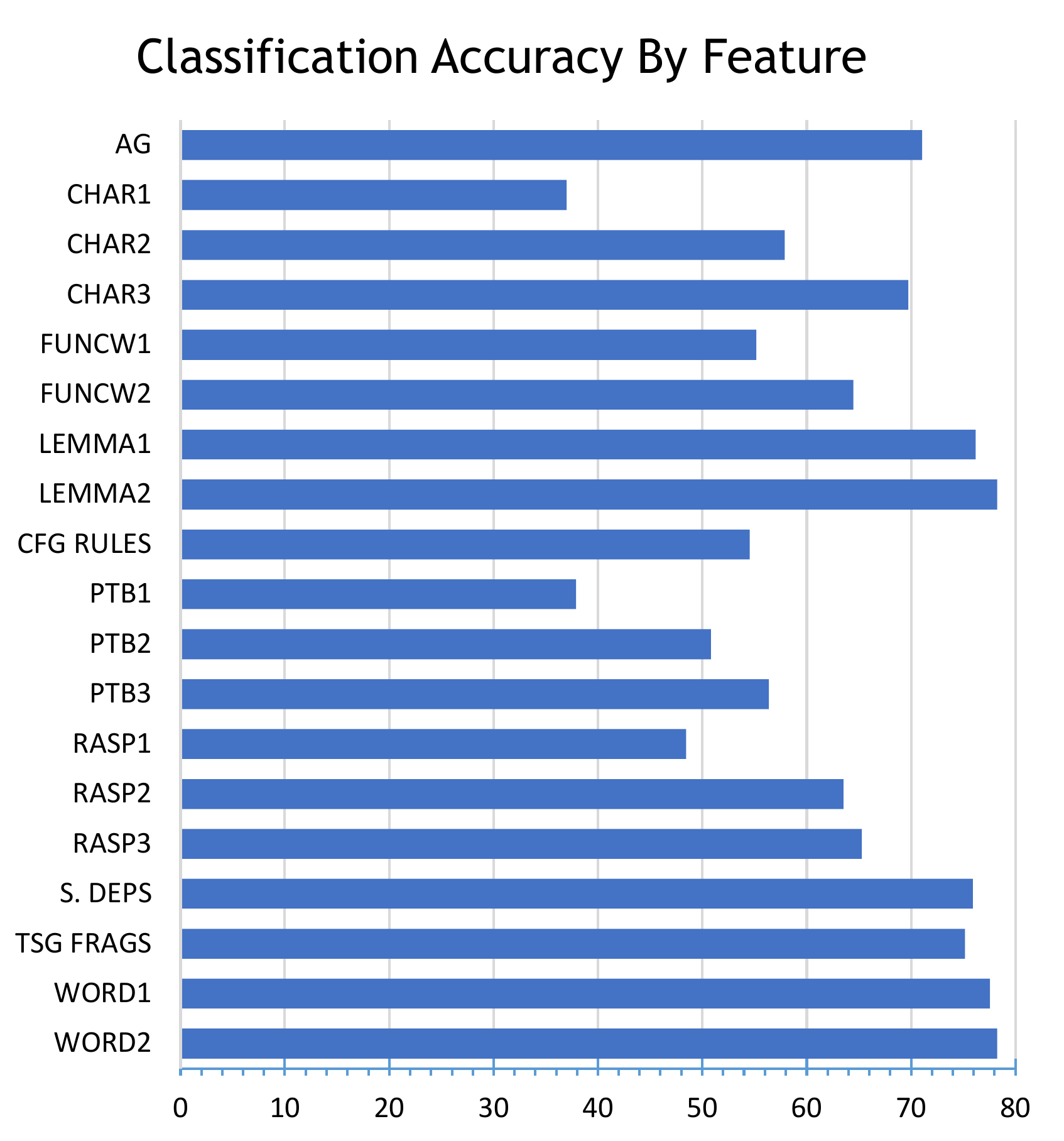}
\caption{NLI accuracy per feature type on the \tfl test set.}
\label{fig:feat-acc}
\end{figure}

As described in \S\ref{sec:models}, we create our ensembles from a set of linear models where each is trained on a different feature type.
We first test our individual classifiers which form this first layer of our three models; this can inform us about their individual performance and the single best feature type.
This is done by training the models on the combined \tfltrain and \tfldev data, which we refer to as \tfltdev, and testing against the \tfltest set.
The results are shown in Figure~\ref{fig:feat-acc}.
We observe that there is a range of performance, and some features achieve similar accuracy. We also note that the best single-model performance is approximately $78\%$. Having shown that our base classifiers achieve good results on their own, we now apply our three ensemble models.

We begin by applying the six ensemble combination methods discussed in \S\ref{sec:ensemble-intro} as part of our first model.
We do this using both cross-validation within \tfltdev and by training our models on \tfltdev and testing on \tfltest. 
The results for all fusion methods, are shown in Table~\ref{tab:ensemble-results}. 

\begin{table}
\caption{Comparing different ensemble classifiers against our baselines and oracles. The CV column lists cross-validation within the \tfltdev data and the Test column is the \tfltest set. Best ensemble result per column in bold.}
\label{tab:ensemble-results}
\centering
\begin{tabular}{clc@{\hskip 1cm}c}
\hline\hline
&\multirow{2}{*}{\textbf{Method}} & \multicolumn{2}{c}{\textbf{Accuracy (\%)}} \\
&\hspace{4cm} 		& CV	  	& Test	\\
\hline
\multirow{5}{*}{\textbf{Baselines}}
&Random Baseline		& ~9\hpt 1 	& ~9\hpt 1 \medskip \\
& 2013 Shared Task Winner	& 84\hpt 5 & 83\hpt 0 \\
& \citet{bykh:2014}				& --- & 84\hpt 8 \\
&\citet{ionescu:2014:nli} 				& 84\hpt 1 & 85\hpt 3 \\
\hline
\multirow{3}{*}{\textbf{Oracles}}
&Oracle				& 96\hpt 1		& 96\hpt 0 \\
&Accuracy@2			& 91\hpt 8		& 92\hpt 0 \\
&Accuracy@3 		& 94\hpt 5 		& 94\hpt 6 \\
\hline
\multirow{6}{*}{\textbf{Ensembles}}
&Plurality Voting	& 82\hpt 6 		& 82\hpt 5\\
&Borda Count			& 81\hpt 2 		& 81\hpt 5 \\
&Mean Probability	& \textbf{82\hpt 6} 		& \textbf{83\hpt 3}\\
&Median Probability	& 82\hpt 4 		& 82\hpt 7\\
&Product Rule		& 80\hpt 3 		& 80\hpt 6 \\
&Highest Confidence	& 80\hpt 1 		& 80\hpt 4 \\
\hline\hline
\end{tabular}
\end{table}

The mean probability combiner which uses continuous values for each class (\S\ref{sec:ens-mean}) achieves the best performance for both of our test sets.
This is followed by the plurality vote and median probability combiners, both of which have similar performance.
Although plurality voting achieves good results, the Borda Count method performs worse.

In our cross-validation experiments, some $2.7\%$ of the instances resulted in voting ties which were broken arbitrarily. This randomness leads to some variance in the results for voting-based fusion methods; running the combiner on the same input can produce slightly different results.

Results from the highest confidence and product rule combiners are the poorest amongst the set, and by a substantial margin. We hypothesize that this is due to fact that they are both highly sensitive to outputs from all classifiers. A single outlier or poor prediction can adversely affect the results. They should generally be used in circumstances where the base classifiers are known to be extremely accurate, which is not the case here. Accordingly, we do not experiment with any further.

These results from this first model comport with previous research reporting that ensembles outperform single-vector approaches (see \S\ref{sec:relwork}); our best ensemble result is some $5\%$ higher than our best feature.

We next apply our meta-classifier (\S\ref{sec:ens-sg}) to both the discrete and continuous outputs generated by the base classifiers.
While the base classifiers remain the same, we train a meta-classifier using each of the machine learning algorithms we listed earlier in \S\ref{sec:learners}.
This results in 15 meta-classification models. Each model is tested using both discrete and continuous input, using both cross-validation and the \tfltest set.
The results for all of these experiments are shown in Table~\ref{tab:metac-results}.

Broadly, we observe two important trends here: the meta-classification results are substantially better than the ensemble combination methods from Table~\ref{tab:ensemble-results}, and that meta-classifiers trained on continuous output perform better than their discrete label counterparts.
This last pattern is not all that surprising since we already observed that the probability-based ensemble combiners outperformed the voting-based combiners. Using continuous values associated with each label provides the meta-learner with more information than a single label, likely helping it make better decisions.

While most of our algorithms perform well, the LDA meta-classifier yields the best results across both input types and test conditions.
These results, $85.2\%$ under cross-validation and $86.8\%$ on \tfltest are already higher than the current state of the art on this data and well exceed the baselines listed in Table~\ref{tab:ensemble-results}.
It is also important to note that the same classifier achieves the best performance across all four testing conditions.

The linear and RBF SVMs also achieve competitive results here.
Instance-based $k$-NN and Nearest Centroid classifiers also do well.
On the other hand, 
decision trees, QDA, and the Perceptron algorithm have the poorest performance across both discrete and continuous inputs.

\begin{table}
\caption{Results from our 15 meta-classifiers applied to \tfl, using both discrete and continuous outputs from the base classifiers.
The best result in each column is in bold. The results can be compared with the baselines in Table~\ref{tab:ensemble-results}.}
\label{tab:metac-results}
\centering
\begin{tabular}{l@{\hskip 1cm}c@{\hskip 0.5cm}c@{\hskip 1.6cm}c@{\hskip 0.5cm}c}
\hline\hline
\multirow{2}{*}{\textbf{Meta-classifier}} & \multicolumn{2}{l}{\textbf{~Discrete}} & \multicolumn{2}{c}{\textbf{Continuous}}\\
\hspace{1cm} 			& CV	  	& Test & CV &  Test\\
\hline
Random Baseline			& ~9\hpt 1 	& ~9\hpt 1 & ~9\hpt 1 & ~9\hpt 1\\
\hline
Linear SVM			& 84\hpt 3 	& 84\hpt 4	& 84\hpt 5 & 85\hpt 2\\
RBF-Kernel SVM		& 84\hpt 2 	& 84\hpt 7	& 84\hpt 6	& 85\hpt 1\\
Logistic Regression	& 83\hpt 9 	& 83\hpt 8	& 84\hpt 3	& 84\hpt 8\\
Ridge Regression	& 84\hpt 3 	& 84\hpt 8	& 84\hpt 2	& 84\hpt 5\\
Perceptron			& 78\hpt 5 	& 81\hpt 5	& 80\hpt 9	& 81\hpt 7\bigskip\\
Decision Tree		& 77\hpt 6 	& 78\hpt 3	& 75\hpt 1	& 75\hpt 2\\
QDA					& 56\hpt 8 	& 57\hpt 3	& 67\hpt 4	& 67\hpt 9\\
LDA					& \textbf{84\hpt 3} 	& \textbf{84\hpt 7}	& \textbf{85\hpt 2}	& \textbf{86\hpt 8} \bigskip\\
Nearest Centroid	& 83\hpt 6 	& 83\hpt 5	& 83\hpt 1	& 83\hpt 5\\
5-NN				& 83\hpt 2 	& 82\hpt 5	& 81\hpt 6	& 82\hpt 0\\
10-NN				& 83\hpt 5 	& 83\hpt 0	& 83\hpt 0	& 84\hpt 5\\
15-NN				& 83\hpt 6 	& 83\hpt 4	& 83\hpt 2	& 84\hpt 3\\
20-NN				& 83\hpt 6 	& 83\hpt 6	& 83\hpt 3	& 84\hpt 7\\
50-NN				& 83\hpt 4 	& 83\hpt 6	& 83\hpt 7	& 84\hpt 2\\
100-NN				& 83\hpt 1 	& 83\hpt 3	& 83\hpt 6	& 84\hpt 1\\
\hline\hline
\end{tabular}
\end{table}

We have thus far shown that ensembles outperform a single-vector approach, and that meta-classifiers achieve state-of-the-art results. Our final \tfl experiment involves applying our hybrid  ensemble of meta-classifiers (\S\ref{sec:ens-mcens}) to determine if we can further improve these results.
Given the results from the previous model, we only test using continuous classifier outputs.

\begin{table}
\caption{Results for our ensembles of meta-classifiers applied to \tfl. Best result per column in bold, best result per row grouping is underlined.}
\label{tab:mcensemble-results}
\centering
\begin{tabular}{llc@{\hskip 1cm}c}
\hline\hline
&\multirow{2}{*}{\textbf{Method}} & \multicolumn{2}{c}{\textbf{Accuracy (\%)}} \\
&\hspace{4cm} 		& CV	  	& Test	\\
\hline
\multirow{5}{3cm}{\textbf{Baselines}} &Random Baseline		& ~9\hpt 1 	& ~9\hpt 1 \medskip \\
& 2013 Shared Task Winner	& 84\hpt 5 & 83\hpt 0 \\
& \citet{bykh:2014}				& --- & 84\hpt 8 \\
&\citet{ionescu:2014:nli} 				& 84\hpt1 & 85\hpt 3 \medskip \\
& Our LDA Meta-classifier (continuous)		& \underline{85\hpt 2} 	& \underline{86\hpt 8}\\
\hline
\multirow{3}{3cm}{\textbf{Decision Tree Ensembles}}
&Random Forest	& \underline{84\hpt 2} 	& \underline{84\hpt 6} \\
&Extra Trees	& 81\hpt 0 	& 82\hpt 7 \\
&AdaBoost		& 75\hpt 3 	& 76\hpt 2 \\
\hline
\multirow{3}{*}{\textbf{Bagging}} & Linear SVM			& 84\hpt 5 		& 85\hpt 2 \\
&Logistic Regression	& 84\hpt 4 		& 84\hpt 9 \\
&Ridge Regression		& 84\hpt 5		& 85\hpt 2 \\
&LDA					& \bf 85\hpt 3		& \bf 87\hpt 1 \\
\hline\hline
\end{tabular}
\end{table}

We experiment with two general methods for creating ensembles of meta-classifiers: boosting and bagging.
Although a single decision tree was not a good meta-classifier, it has been shown that ensembles of trees can perform very well  \cite{banfield2007comparison}.
We experiment with random forests, extra trees and AdaBoost for creating such tree-based ensembles.
We also apply bagging to several of the best meta-classifiers from Table~\ref{tab:metac-results}: SVMs, Logistic Regression, Ridge Regression and LDA.
To combine the ensemble of meta-classifiers we use the mean probability combiner, given its better performance among the combiners listed in Table~\ref{tab:ensemble-results}.

Results from these models are shown in Table~\ref{tab:mcensemble-results}.
As expected, we observe that the tree-based methods receive a substantial performance increase compared to the single decision tree meta-classifier from the previous model. Random forests provide the biggest boost, improving performance by almost $10\%$.
However, this is still lower than our LDA meta-classifier.

Applying bagging to our discriminative meta-classifiers, we observe that we gain a small improvement over the previous model. The LDA-based method again outperforms the others, and while the improvement is not huge, it sets a new upper bound for \tfl. In fact, this result is only $9\%$ lower than the oracle accuracy of~$96\%$.

In designing this setup we were initially concerned that the addition of further layers could lead to the addition of errors in the deeper classifiers, resulting in performance degradation.
However, this was not the case and accuracy increased, if only slightly.

\begin{figure}
\centering
\includegraphics[trim=0 0 0 0,clip,width=0.8\textwidth]{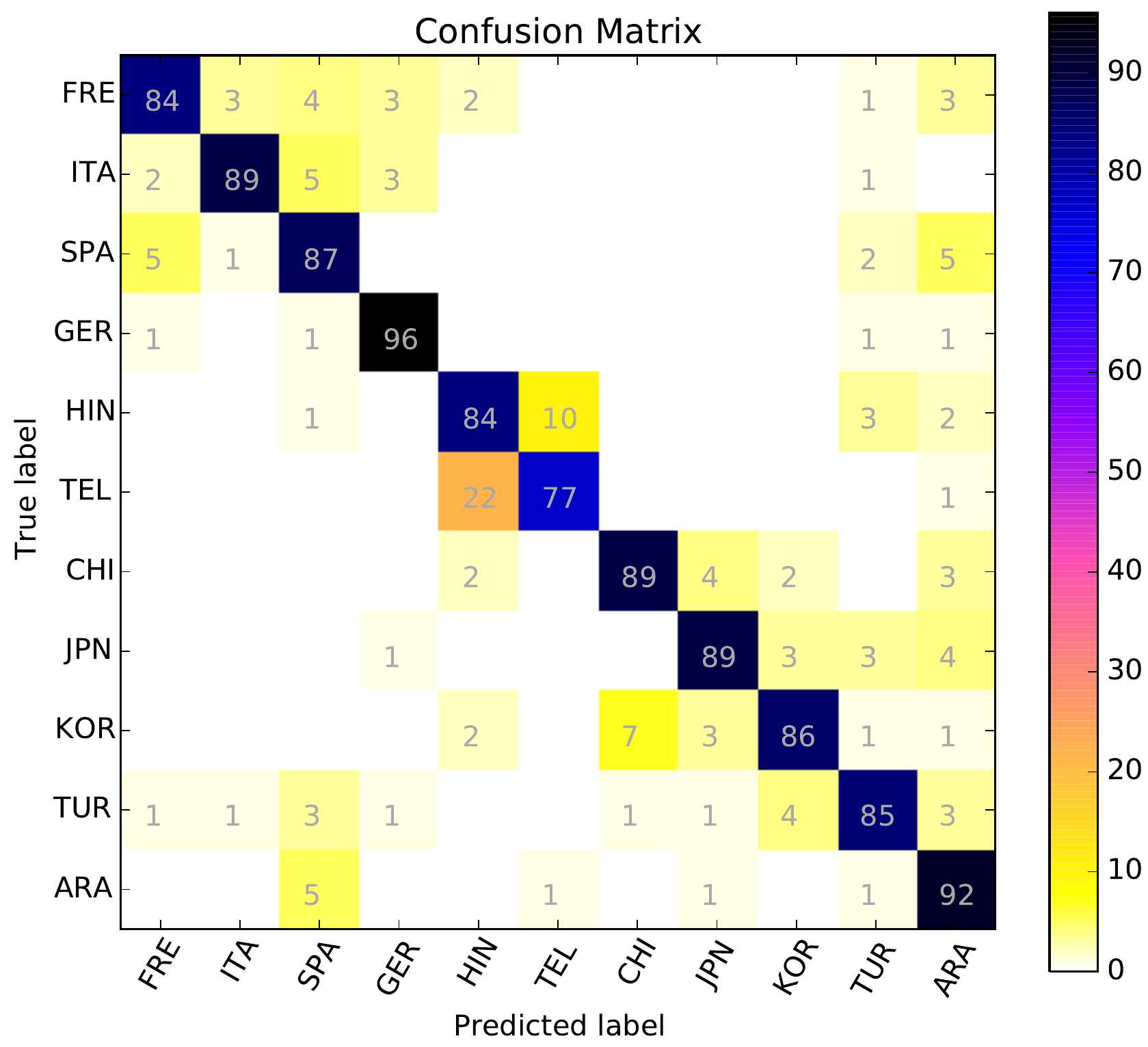}
\caption{Confusion matrix of our best performing NLI system, achieving 87.1\% accuracy on the \tfltest set.}
\label{fig:t11confm}
\end{figure}

A confusion matrix of our best system's predictions on the \tfltest set is presented in Figure~\ref{fig:t11confm}.
The labels in the matrix have been ordered in a way similar to Figure~\ref{fig:TOEFL11-languages} in order to group similar languages together.
We achieve our best performance on German texts, with only 4 misclassifications. In the top left corner we also observe some confusion between the romance languages.
We also observe the asymmetric confusion between Hindi and Telugu, as discussed in previous research \cite{malmasi-tetreault-dras:2015}.
Another interesting observation is that Arabic, which has poor precision, receives misclassifications from every other class, except Italian. This trend can be observed in the last column of the matrix.

\begin{table}
\caption{Per-class performance breakdown of our top system's results on the \tfltest set. Best result per column in bold. Best and worst performances per column as also highlighted in green/yellow.}
\label{tab:tfl-breakdown}
\center
\begin{tabular}{rccc}
\hline\hline
Class	& Precision &  Recall	& F1-score \\ 
\hline
\tt ARA	&	0.80	&	0.92	&	0.86 \\
\tt CHI	&	0.92	&	0.89	&	0.90 \\
\tt FRE	&	0.90	&	0.84	&	0.87 \\
\tt GER	&	0.92	&	\bf \colorbox{green}{0.96}	&	 \bf \colorbox{green}{0.94} \\
\tt HIN	&	\colorbox{yellow}{0.75}	&	0.84	&	\colorbox{yellow}{0.79} \\
\tt ITA	& \bf \colorbox{green}{0.95}	&	0.89	&	0.92 \\
\tt JPN	&	0.91	&	0.89	&	0.90 \\
\tt KOR	&	0.91	&	0.86	&	0.88 \\
\tt SPA	&	0.82	&	0.87	&	0.84 \\
\tt TEL	&	0.88	&	\colorbox{yellow}{0.77}	&	0.82 \\
\tt TUR	&	0.87	&	0.85	&	0.86 \\
\hline
Average	&	0.87	&	0.87	&	0.87 \\
\hline\hline
\end{tabular}
\end{table}

We also assessed per-class performance using precision, recall and the F1-score, with results listed in Table~\ref{tab:tfl-breakdown}.
As shown in the confusion matrix, Hindi and Telugu have the worst performance.
Recalculating the values without those two classes, the average F1-score improves to $0.89$.

\subsection{Experiments On Other Languages}
\label{sec:exp2}

The second set of our experiments focus on investigating the generalizability of our findings so far. The result patterns observed on \tfl have been stable across the training and test set, but we now apply them to other datasets to assess their generalizability on different languages and data sources.

The experiments in this section are conducted on the Chinese and Norwegian datasets described in \S\ref{sec:data}.
As these datasets do not have a predefined test set like \tfl, these experiments were performed using stratified cross-validation, as discussed in \S\ref{sec:evaluation}.
Previous experiments on these corpora have also been conducted using cross-validation only.

\begin{table}[!h]
\caption{Results for our three models on the Chinese and Norwegian datasets, using continuous classifier outputs. Best result per column in bold, best result per row grouping is underlined.}
\label{tab:otherl2-results}
\centering
\begin{tabular}{llc@{\hskip 1cm}c}
\hline\hline
&\multirow{2}{*}{\textbf{Feature}} & \multicolumn{2}{c}{\textbf{Accuracy (\%)}} \\
&\hspace{4cm} 		& Chinese	  	& Norwegian	\\
\hline
\multirow{3}{*}{\textbf{Baselines}}
&Random Baseline			& ~9\hpt 1 	& 10\hpt 0\\
&Majority Class Baseline	& 12\hpt 9 	& 13\hpt 0\\
&Current Best Result		& 70\hpt 6 	& 78\hpt 6 \\
\hline
\multirow{3}{*}{\textbf{Oracles}}
&Oracle				& 92\hpt 2		& 94\hpt 5 \\
&Accuracy@2			& 76\hpt 9		& 89\hpt 3 \\
&Accuracy@3 		& 84\hpt 7 		& 92\hpt 5 \\
\hline
\multirow{4}{*}{\textbf{Ensembles}} &
Plurality Voting	& 68\hpt 5 		& 75\hpt 7 \\
&Borda Count		& 66\hpt 4 		& 76\hpt 7 \\
&Mean Probability	& \underline{71\hpt 1} 		& \underline{77\hpt 9} \\
&Median Probability	& 66\hpt 1 		& 77\hpt 3 \\
\hline
\multirow{4}{2.9cm}{\textbf{Meta-classifier}}
&Linear SVM				& 75\hpt 4 		& 78\hpt 6 \\
&Logistic Regression	& 74\hpt 6 		& 79\hpt 6 \\
&Ridge Regression		& 71\hpt 4		& 78\hpt 6 \\
&LDA					& \underline{75\hpt 9}		& \underline{81\hpt 0} \\
\hline
\multirow{4}{2.9cm}{\textbf{Meta-classifier Bagging}} &Linear SVM				& 75\hpt 5 		& 78\hpt 7 \\
&Logistic Regression	& 75\hpt 1 		& 80\hpt 1 \\
&Ridge Regression		& 71\hpt 4		& 78\hpt 8 \\
&LDA					& \textbf{76\hpt 5}		& \textbf{81\hpt 8} \\
\hline\hline
\end{tabular}
\end{table}

We utilize the top performing models tested in \S\ref{sec:exp1}:
four ensemble combiners, four bagging-based meta-classifiers and four ensembles of meta-classifiers.
These selected models, and their results, are listed in Table~\ref{tab:otherl2-results}.

The oracle values for both datasets are quite high at over $90\%$, similar to \tfl (which was listed in Table~\ref{tab:ensemble-results}).
The ensemble model does well, beating the previously reported best result for Chinese and coming close for Norwegian.
Just like our previous experiments, the mean probability combiner yields the best performance.

The meta-classifier model achieves a new state of the art for both datasets, just as it did for \tfl.
Also consistent with the previous experiment, LDA achieves the best results for both datasets.
Finally, the ensemble of meta-classifiers yields additional improvement over the single meta-classifier model, achieving our best results. We achieve $76.5\%$ accuracy on the Chinese data and $81.8\%$ on the Norwegian data, both substantial improvements over previous work.
These results show that these classification models are applicable to other datasets.
The results followed the same pattern across all three datasets, with LDA-based meta-classification yielding top results.

\subsection{Statistical Significance Testing}
\label{sec:exp-mcnemar}

An important question that arises in various contexts within machine learning deals with determining which methods outperform others on a given problem \citep{dietterich1998approximate}. Such questions are often addressed using statistical significance testing, which can help base such research and claims about results in a rigorous empirical foundation.

However, this trend has not been adopted within NLI. Most publications have reported cross-validation accuracy, or more recently, accuracy on the \tfltest set. Given the increasing accuracies reported by recent research, we believe that the use of statistical tests can be very beneficial to future NLI research.

Reliable statistical tests for comparing two classifiers require the availability of a common test set, a need which has been recently addressed by the \tfltest set.\footnote{This remains an issue for other datasets where no test set exists and only cross-validation results are usually reported, as we noted earlier.}
Since we are then evaluating the classifier outputs for the same samples, a test for paired nominal data is suitable.
McNemar's test \citep{mcnemar1947note} is a non-parametric method to test for significant differences in proportions for paired nominal data.
In the context of machine learning it is often used to compare the performance of distinct algorithms on the same data \citep{dietterich1998approximate} as it does not assume independent samples and has a low Type I error rate \citep{furnkranz2002round}.
It is the most commonly used for pairwise classifier comparison 
and has been used in a wide range of machine learning applications \cite{west2000neural,aue2005customizing}.

This is also the test we propose for use within NLI.
In this section we briefly describe the test and demonstrate its application for NLI.
The interested reader can find more details about methods for evaluating the statistical significance of classifier differences in the work of \citet{foody2004thematic}.

McNemar's test is a non-parametric method based on creating a 2$\times$2 contingency table for the outcomes of a pair of tests (classifiers in our case), tabulating the number of instances where their predictions agree or disagree. The row and column marginals are calculated, and a test statistic is then used to determine if the marginal probabilities for each classifier are the same.
An example of such a contingency table for two classifiers $C_{a}$ and $C_{b}$ is given below in Table~\ref{tab:mcnemar-contingency}.

\begin{table}[!h]
\caption{Example contingency table for the outputs of two classifiers. The four cells represent the number of concordant and discordant classifications and misclassifications between the two methods.}
\label{tab:mcnemar-contingency}
\centering
\scalebox{0.75}{
\begin{tabular}{lcc}
& \textbf{$C_{b}$ Correct} & \textbf{$C_{b}$ Wrong} \\
\textbf{$C_{a}$ Correct} & $n_{11}$ & $n_{10}$ \\
\textbf{$C_{a}$ Wrong}   & $n_{01}$ & $n_{00}$ \\
\end{tabular}
}
\end{table}

The four table cells represent the number of concordant and discordant classifications and misclassifications between the two methods.
The null hypothesis states that both classifiers have equal error rates ($n_{01} = n_{10}$, \ie the discordant predictions are the same)\footnote{This is sometimes stated as $\frac{n_{01}}{n_{01} + n_{10}} = 0.5$} and the alternative hypothesis is that the error rates differ.
The test statistic is based on a chi-square distribution, with additional continuity correction to account for the fact that a continuous distribution is being used to represent a discrete one \citep{dietterich1998approximate,foody2004thematic}.
The test statistic is given in Eq.~\ref{eq:mcnemar}.

\begin{equation}
\label{eq:mcnemar}
\chi = \frac{{(|n_{01} - n_{10}| - 1)}^{2}}{n_{01} + n_{10}}
\end{equation}
\\

Having defined the test, the final requirement for its use is the availability of predictions from different systems.
However, this is an important factor that has hindered the adoption of statistical tests in NLI.
Despite having a predefined test set, an obstacle here is that although the \tfl corpus has been used in most NLI work since the 2013 shared task, most researchers do not make their predictions available.
The availability of these predictions enables the application of statistical significance testing for classifier evaluation.
The work of \citet{malmasi-tetreault-dras:2015} was an initial step in this direction by making available all $144$ submissions from the 2013 shared task, including that of the winning system.\footnote{Available from \url{http://web.science.mq.edu.au/~smalmasi/resources/nli2013}}
As part of this work, we also make available the predictions of our best system on the \tfltest set.
During the course of this research, \citet{ionescu:2014:nli} also provided us with the predictions from their state-of-the-art system, which we also make available.\footnote{These two sets of predictions are available at\\ \url{http://web.science.mq.edu.au/~smalmasi/resources/nli-predictions}}
The availability of this data can become increasingly important as state-of-the-art results move closer towards the oracle upper bounds.

\begin{table}[!ht]
\caption{Results for statistical significance testing between our top system and two previous state-of-the-art systems. The \textit{p} values from McNemar's test are reported. * = significant at the 0.001 level, ** = significant at the 0.01 level and ***~=~significant at the 0.05 level.}
\label{tab:mcnemar}
\centering
\begin{tabular}{rlll}
& \bf Jarvis et al. & \bf Ionescu et al. & \bf Our Method \\
\hline
\bf \citet{jarvis-bestgen-pepper:2013:BEA8} & --- &  $0.1082$ & $0.0001$* \\
\bf \citet{ionescu:2014:nli} &  -- &  --- &  $ 0.0314$*** \\
\bf Our Method & -- & -- & --- \\
\end{tabular}
\end{table}

We now evaluate the performance of our top model against that of the two previous state-of-the-art systems which were used as baselines in our experiments, using the aforementioned prediction data.
We report the pairwise \textit{p} values for the test, as listed in Table~\ref{tab:mcnemar}.
They show that the improvement in our results is significantly better than both of the baselines.
In contrast, they also show that the results of \citet{ionescu:2014:nli} were not significantly better than the previous best result.
This analysis highlights the utility of using such statistical tests for NLI.

\section{Discussion}
\label{sec:discussion}

We presented the first comprehensive study of meta-classification techniques for NLI, achieving state-of-the-art accuracy on three major datasets for the task.
This is the most comprehensive and systematic application of multiple classifier systems for NLI, evaluating three types of increasingly sophisticated classification models.

We applied many different methods from the armamentarium of machine learning algorithms, and the observed consistency was an important facet of our results. The performance patterns of our models were similar across different languages and dataset, with the same model configurations achieving the best results across different test sets and corpora.
This differs to the work of \citet{ionescu:2014:nli}, where their best results on different sets were achieved using different parameters, or that of \citet{bykh:2014}, who did not test their method on different datasets.

The application of these methods is not limited to cross-validation studies and we have attempted to apply them elsewhere. 
During the course of developing these methods we evaluated them under test conditions by using them to compete in several shared tasks in different tasks.
Although a detailed exposition exceeds the scope of the present work, we briefly mention our results.
The ensemble classifier was used to participate in the 2015 Discriminating Similar Language shared task, and was the winning entry among the 10 participating teams.
The ensemble was also used to train a system to participate in the Complex Word Identification task at SemEval 2016 (Track 11), with our systems ranking in second and third place.
Finally, the meta-classifier ensemble approach described here was the basis of an entry in the 2016 Computational Linguistics and Clinical Psychology (CLPsych) shared task, where it also ranked in first place among 60 systems.
We believe that these results, in conjunction with the state-of-the-art NLI performance reported in the present paper, highlight the utility of the classification models we described here for various NLP tasks.

We also introduced the possibility of statistical significance testing within NLI, making available two new sets of predictions to facilitate this.
Work in NLI has not yet begun to use such statistical significance testing for comparing results, although this is something that becomes increasingly desirable as the 
relative differences between proposed methods begin to narrow and results get closer to the oracle upper bound.
Although the predictions needed for such analyses are currently only available for \tfl, we hope that their use will be adopted for future NLI work using other datasets.

Future work can be directed towards answering some of the following questions.
\textit{Why does LDA outperform other meta-classification methods?}
In-depth examination of the trained models -- something beyond the scope of this work -- may reveal interesting clues about what the model is learning. This knowledge could possibly help improve meta-classifier feature engineering.

\textit{How does the amount of training data affect meta-classifier performance?}
This analysis, along with further evaluation of the models' learning curves, could inform us about training data requirements as well as bias-variance and overfitting issues.

\textit{Can meta-classifiers improve cross-corpus performance?}
As additional datasets suitable for NLI become available, this has enabled the application of cross-corpus evaluation to assess how well the methods generalize across data from different genres and sources \citep{malmasi-dras:2015:nli}.
The meta-classifier approach has yet to be tested in such a scenario and future experiments in this context could provide insightful results.

\clearpage
\bibliographystyle{plainnat}
\bibliography{nlisg}

\end{document}